\documentclass[11pt]{article}

\usepackage[preprint]{acl}

\usepackage{times}
\usepackage{latexsym}

\usepackage[T1]{fontenc}

\usepackage[utf8]{inputenc}

\usepackage{microtype}

\usepackage{inconsolata}

\usepackage{graphicx}

\usepackage{xcolor}
\usepackage{booktabs}
\usepackage{tabularx}
\usepackage{amsmath}
\usepackage{amssymb}
\usepackage{multirow}
\usepackage{adjustbox}
\usepackage{colortbl}
\usepackage{enumitem}
\usepackage{listings}
\usepackage{xcolor}
\usepackage[most]{tcolorbox}
\newcommand{\eat}[1]{}
\usepackage{amsmath}
\newcount\mycount
\newcommand{\insertEmptyLines}[1]{%
  \mycount=#1%
  \loop
    \ifnum\mycount>0
       \\
       \advance\mycount by -1
  \repeat
}

\definecolor{structural}{HTML}{DFF2E1}
\definecolor{conceptual}{HTML}{F4D7ED}
\definecolor{property}{HTML}{FDE4D0}
\definecolor{ranking}{HTML}{FFF2CC}
\definecolor{advanced}{HTML}{D6E9F8}
\definecolor{design}{HTML}{E3CED4}

\newcommand{\qwen}[1]{\texttt{Qwen-2.5-7B}}
\newcommand{\qwenbig}[1]{\texttt{Qwen-2.5-14B}}
\newcommand{\qwenbigger}[1]{\texttt{Qwen-2.5-32B}}
\newcommand{\microsoftphi}[1]{\texttt{Phi-4-14B}}
\newcommand{\gptOSS}[1]{\texttt{GPT-OSS-20B}}

\newcommand{\dataset}[1]{\texttt{PolyBench}}
\newcommand{\datasetTrain}[1]{\texttt{PolyBench$_{train}$}}
\newcommand{\datasetTest}[1]{\texttt{PolyBench$_{test}$}}
\newcommand{\Llamole}[1]{\texttt{Llamole-8B}}
\newcommand{\LlamaSmol}[1]{\texttt{LlaSMol-8B}}
\newcommand{\etherzero}[1]{\texttt{ether0-24B}}
\newcommand{\ChemLLM}[1]{\texttt{ChemLLM-7B}}

\newcommand{\gptFourOMini}[1]{\texttt{GPT-4o-Mini}}
\newcommand{\gptFourO}[1]{\texttt{GPT-4o}}
\newcommand{\claudeSonnet}[1]{\texttt{Claude-3.5-Sonnet}}
\newcommand{\claudeSonnetfoursix}[1]{\texttt{Sonnet-4.6}}

\newcommand{\sonnet}[1]{\texttt{Sonnet}}
\newcommand{\fourOMini}[1]{\texttt{4o-Mini}}
\newcommand{\gptfive}[1]{\texttt{GPT-5}}
\newcommand{\gptfivefour}[1]{\texttt{GPT-5.4}}

\newcommand{\taskOnestrunderstanding}[1]{\texttt{Structural Understanding}}
\newcommand{\taskTwopolymerconcepts}[1]{\texttt{Polymer Concepts}}
\newcommand{\taskThreeproppred}[1]{\texttt{Property Prediction}}
\newcommand{\taskFourpropcomprank}[1]{\texttt{Property Comparison \& Ranking}}
\newcommand{\taskFivereasoning}[1]{\texttt{Advanced Property Reasoning}}
\newcommand{\taskSixdesign}[1]{\texttt{Design \& Synthesis}}

\newcommand{\smiles}[1]{\texttt{SMILES}}

\newcommand{\squishlist}{
  \begin{list}{$\bullet$}
    { \setlength{\itemsep}{0pt}      \setlength{\parsep}{3pt}
      \setlength{\topsep}{3pt}       \setlength{\partopsep}{0pt}
      \setlength{\leftmargin}{1.5em} \setlength{\labelwidth}{1em}
      \setlength{\labelsep}{0.5em} } }
\newcommand{\reallysquishlist}{
  \begin{list}{$\bullet$}
    { \setlength{\itemsep}{0pt}    \setlength{\parsep}{0pt}
      \setlength{\topsep}{0pt}     \setlength{\partopsep}{0pt}
      \setlength{\leftmargin}{0.2em} \setlength{\labelwidth}{0.2em}
      \setlength{\labelsep}{0.2em} } }

 \newcommand{\squishend}{
     \end{list} 
 }

\title{Teaching and Evaluating LLMs to Reason About\\ Polymer Design Related Tasks}

\author{
  \textbf{Dikshya Mohanty},
  \textbf{Mohammad Saqib Hasan},
  \textbf{Syed Mostofa Monsur},
  \textbf{Size Zheng},
\\
  \textbf{Benjamin Hsiao},
  \textbf{Niranjan Balasubramanian}
\\[6pt]
  Stony Brook University
\\
  \texttt{dimohanty@cs.stonybrook.edu}
}

\begin{document}
\maketitle
\begin{abstract}
Research in \textit{AI4Science} has shown promise in many science applications, including polymer design. However, current LLMs are ineffective in this problem space because: (i) most models lack polymer-specific knowledge, and (ii) existing aligned models have limited coverage of knowledge and capabilities relevant to polymer design. Addressing this, we introduce \dataset{}, a large-scale training and test benchmark dataset of more than 125K polymer design-related tasks, leveraging a knowledge base of more than 13 million data points obtained from experimental and synthetic data sources to ensure broad coverage of polymers and their properties. For effective alignment using \dataset{}, we introduce a knowledge-augmented reasoning distillation method that augments this dataset with structured CoT. Furthermore, tasks in \dataset{} are organized from simple to complex analytical reasoning problems, enabling generalization tests and diagnostic probes across the problem space. Experiments show that small- \& mid- sized language models (SLMs) with 7B to 32B parameters, trained on \dataset{}, outperform similar-sized models and remain competitive with closed-source frontier LLMs on \dataset{}'s test dataset, while demonstrating performance gains on external polymer benchmarks. 
\end{abstract}

\section{Introduction}

Polymer design and synthesis is a complex and resource-intensive endeavor involving multiple steps: selecting constituent monomers (smaller compound units)~\cite{Pillai01012010,polym13183147,jablonka2021bias,Tran_2023,ALDAWOOD2025114369}, designing the procedure for synthesis~\cite{ONORATI2026118599, doi:10.1021/ie950072c, molecules30153244}, and validating whether the proposed polymer adheres to desired structure and properties~\cite{doi:10.1021/acs.macromol.1c02597,sahu2022polymer, polymetrix}.
The search space of experimental parameters and possible polymer candidates to investigate becomes exponential during synthesis, making laboratory studies costly and, in certain cases, infeasible.

\begin{figure}
\centering
\includegraphics[width=0.95\linewidth, keepaspectratio]{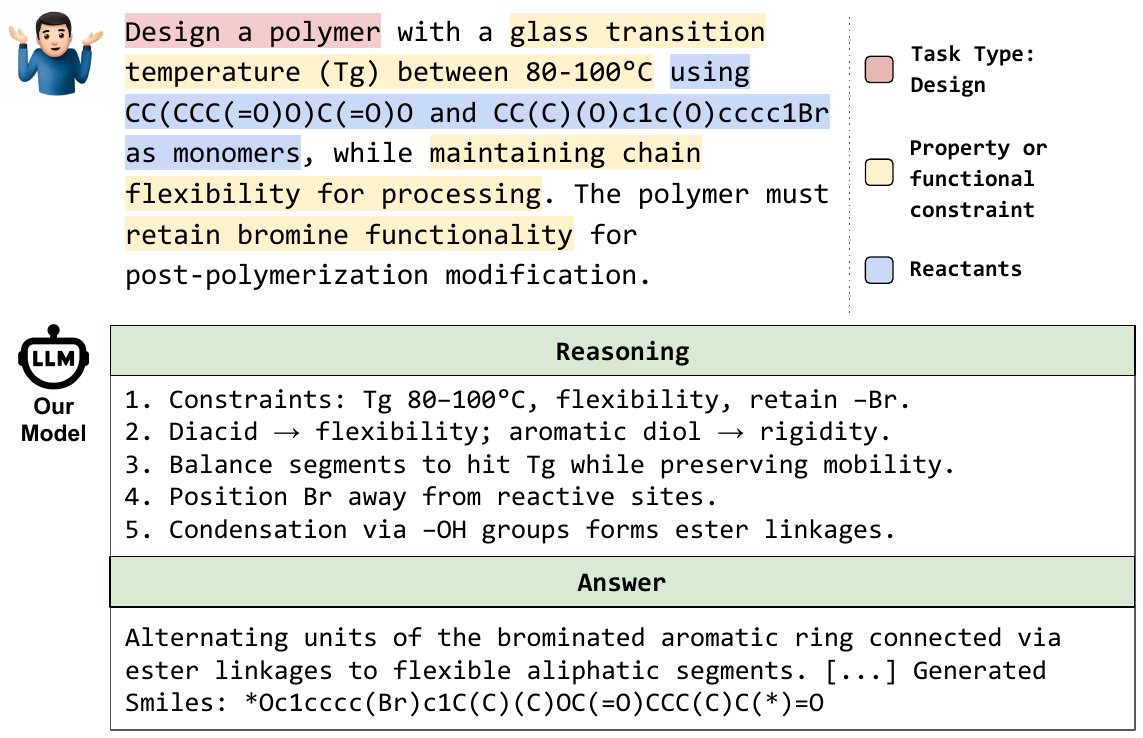}
\caption{\textbf{Example from \dataset{}.} A polymer design prompt with color-coded constraints spanning \texttt{property targets}, \texttt{reactants}, and \texttt{functional constraints}. Solving it requires jointly satisfying all three, illustrating the \textbf{multi-constraint nature} of polymer design. \dataset{}-trained models decompose the prompt into step-by-step reasoning and produce a valid SMILES repeat unit that satisfies every constraint.}
\label{fig:main-figure}
\vspace{-1.5em}
\end{figure}



Large Language Models (LLMs) can be used to support a wide range of information needs that polymer scientists may have during the design and synthesis phases.  
Compared to traditional materials science informatics approaches which develop fragmented solutions that target individual or small subsets of information needs~\cite{queen_polymer_2023, guo-etal-2022-multilingual, xu_transpolymer_2023, kuenneth_polybert_2023}, LLMs can support natural language interaction and reasoning for a wide range of complex, multi-constraint design tasks (see Figure~\ref{fig:main-figure} for an example). 


\noindent Serving such polymer design-related information needs presents multiple challenges: handling complex hierarchical syntactic structures with repeating and branching units (via SMILES\footnote{Simplified Molecular Input Line Entry System: a standard text-based notation that encodes molecular structure as a linear string of atoms and bonds.}), and connecting how structure relates to properties i.e., understanding how specific "code" segments (monomers) dictate bulk physical behavior (properties). 

\noindent Existing LLM-based solutions, however, are ineffective for polymer design-related tasks due to several key limitations: First, as we show later, due to a lack of domain alignment, current frontier LLMs \cite{singh2025openaigpt5card,openai2024gpt4o,openai2024gpt4omini,anthropic2024claude35sonnet} and even chemistry-aligned LLMs \cite{ether0, chemllm, llasmol} fare poorly. Second, there is a lack of multi-objective reasoning tasks and tests of generalization; most polymer-specific training focuses on a narrow set of atomic polymer tasks \cite{xu_transpolymer_2023, kuenneth_polybert_2023, qiu_-demand_2024, zhang_multimodal_2025, zhou_polycl_2025} and fails to reason across multiple properties, constraints, and goals simultaneously or support the wide range of information needs such as filtering, ranking, and synthesis planning. 


\noindent We need large-scale broad-coverage datasets for developing and evaluating effective LLM-based solutions for polymer design. However, constructing such a resource is difficult because (i) the dataset must be grounded in verified scientific outcomes with precise syntactic structural representations, (ii) multi-dimensional design-centric data is often sparse and finding complete records across different dimensions is difficult, and (iii) there are no ready sources that provide reasoning traces that connect the outcome data points to the tasks themselves. We address these by designing a systematic, subject matter expert (SME) informed, and comprehensive process with four key contributions:
\vspace{-0.5em}
\paragraph{1) Design-Related Tasks Grounded in Scientific Data} Using existing polymer data sources of experimental and other fundamental knowledge, along with chemistry tools such as RDKit \cite{rdkit}, we curate polymer profiles and derive \dataset{}, a natural language benchmark of $125k+$ polymer design-related tasks, with inputs from polymer scientists.
\vspace{-0.5em}
\paragraph{2) Task Design and Organization} To reflect the multi-objective nature of polymer design, we need broad coverage of tasks, reasoning patterns, and the kinds of information needs relevant to polymer design. Keeping this as our goal, we organize \dataset{} to cover six different types of \emph{task categories} spanning across basic understanding to complex design---structural understanding, conceptual knowledge, property prediction, property comparison and ranking, advanced reasoning covering complex analytical problems, as well as design and synthesis tasks. 
\vspace{-0.5em}
\paragraph{3) Knowledge-Augmented Reasoning and Distillation} To teach models effective step-by-step reasoning grounded in complete structural and property data, which mirrors how SMEs navigate such tasks, we devise a data-grounded distillation framework by prompting frontier models with the correct background information (such as polymer profiles, reaction details) and reasoning strategy outlines informed by SMEs. We further curate the reasoning traces thus distilled through automated fact-checking for additional verification.
\vspace{-0.5em}
\paragraph{4) Benchmarking and Understanding gaps in model reasoning} Our evaluations show that frontier models and even domain-aligned chemistry LLMs perform poorly on design and synthesis tasks. Training small LMs (7B--14B) using our knowledge-augmented reasoning distillation procedure leads to substantial improvements, especially in design and synthesis tasks. Finally, using the diagnostic supports in our benchmark, we conduct an error analysis that helps understand whether model failures stem from a lack of foundational chemical knowledge (a skill gap) or the inability to combine these foundational skills in compositional multi-step reasoning (a compositionality gap). 


\section{Related Work}
\paragraph{Benchmarks for Chemistry Reasoning.} While science QA benchmarks like GPQA \cite{Rein2023GPQAAG} exist, they have limited chemistry coverage. Hence, chemistry-specific benchmarks exist: \textsc{ChemBench} \cite{Mirza2024AreLL}, \textsc{ChemLLMBench}~\cite{guo2023chemllmbench} and \textsc{ChemCoTBench} \cite{li2025beyond} evaluate LLMs on general chemistry tasks, with the latter incorporating chain-of-thought supervision; and \textsc{PolyCompQA} \cite{circi2024extracting} targets compositional property prediction and question answering, while \textsc{PolyIE} \cite{cheung-etal-2024-polyie} focuses on information extraction from polymer literature. PolyInfo~\cite{polyinfo} is a polymer-properties database which includes structure, synthesis, processing and experimental properties data. PI1M~\cite{PI1M} is a large-scale synthetic polymer dataset. Recently,~\cite{liu2026polyrealbenchmarkrealworldpolymer} introduced PolyReal - a multimodal polymer benchmark dataset focused on end-to-end polymer experimentation workflows. Existing benchmarks are generally limited to small molecules and lack task diversity. Also, using polymer-specific resources, like PolyInfo~\cite{polyinfo} is restricted due to licensing issues. Hence, no broad benchmark with diverse polymer reasoning tasks, such as structural understanding and multi-property inference, exist. We address this by creating \dataset{}, with large coverage of polymer reasoning tasks. \\
\vspace{-1pt}
\textbf{LLMs for Chemistry.} Multiple chemistry LLMs exist: \textsc{Ether0}~\cite{ether0} is a domain-aligned Mistral-24B reasoning model; \textsc{ChemLLM}~\cite{chemllm} is a 7B InternLM2 instruction-tuned on general chemistry tasks; \textsc{LLaSMol}~\citep{llasmol} is an 8B LLaMA designed for molecular question-answering; \textsc{Llamole}~\citep{llamaole} is a multimodal model for retrosynthesis planning; \textsc{StructChem}~\cite{ouyang_structured_2024} incorporates reasoning steps for chemical structure understanding; and \textsc{RL-Guider}~\cite{liu-etal-2025-rl} applies RLHF to guide molecular optimization. These LLMs, however, generalize for small molecules and not for polymers. \\
\vspace{-1pt}
\noindent \textbf{LLMs for Polymer Tasks.} Researchers have created polymer-specific LLMs: \textsc{TransPolymer}~\cite{xu_transpolymer_2023} and \textsc{PolyBERT}~\cite{kuenneth_polybert_2023} use transformer-based models for property learning from SMILES; \textsc{PolyLLMEM}~\cite{zhang_multimodal_2025} uses multi-modal features for property prediction; \textsc{PolyTao}~\cite{qiu_-demand_2024} uses transformer-based models for property-to-polymer design; \textsc{PolyCL}~\cite{zhou_polycl_2025} is a transformer trained with contrastive learning for property prediction; and \citet{qiu_introducing_2025} introduce \textsc{PolySea}, an LLM agent for general polymer questions including property prediction and inverse design. More recently,~\cite{polyt5} introduce \textsc{PolyT5} -  an encoder-decoder foundation chemical model, and ~\cite{nigam2026polymeragentlargelanguagemodel} introduce \textsc{PolyAgent} - a LLM-guided, closed loop framework - both designed for polymer-related tasks like property prediction and design. \\
\vspace{-2pt}
However, these polymer LLMs focus on a narrow range of tasks emphasizing property prediction and generation, with limited generalizability to out-of-distribution settings, and do not incorporate \texttt{CoT} or compositionality. This gap motivates \dataset{}, one of the first datasets for polymer domain alignment with \texttt{CoT} supervision, encompassing multiple polymer-related tasks for robust evaluation of generalization.
\vspace{-2pt}
\eat{
\begin{figure*}[t!]
    \centering
    \includegraphics[width=0.45\textwidth, angle=90]{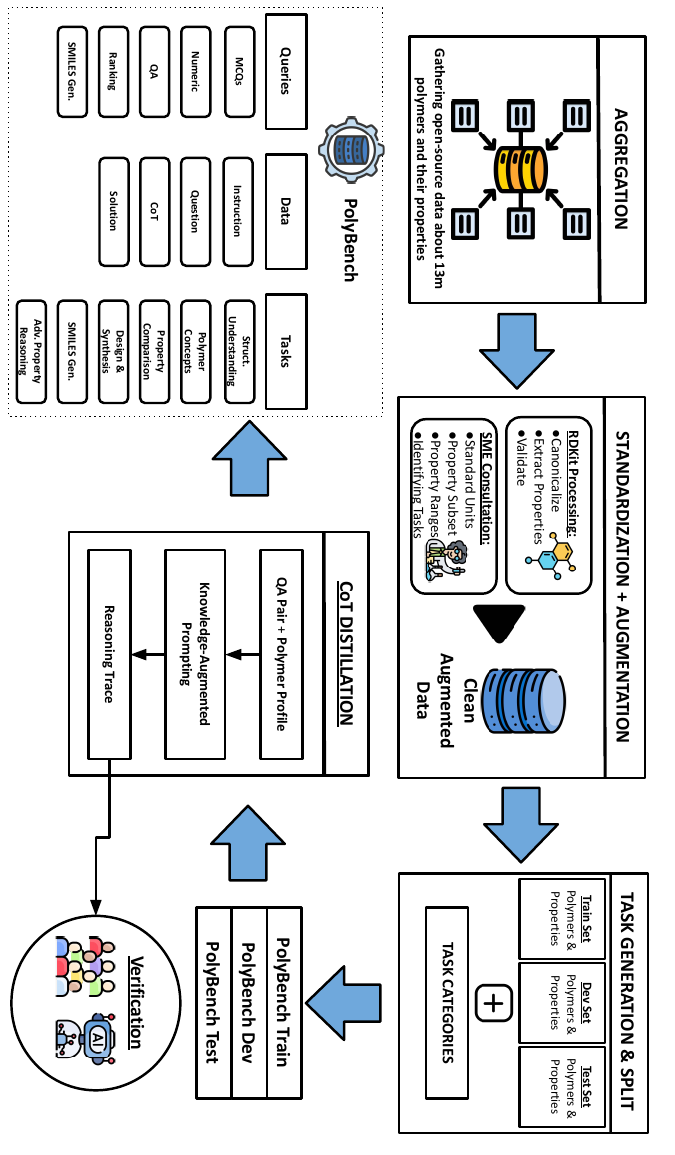}
    \caption{\textbf{PolyData Creation Pipeline}. We aggregate data from open-source databases covering $13$m polymers, standardize notations and properties based on SMEs' inputs, and augment with \texttt{RDKit}-computed features. The data is split into train/dev/test sets with non-overlapping polymers to ensure out-of-distribution evaluation. Each instance undergoes chain-of-thought distillation where polymer profiles are provided to teacher models via \textit{knowledge-augmented prompting}. Generated reasoning traces are verified through automated judgment and manual review, producing \texttt{(instruction,question,CoT,solution)} tuples across six task categories.}
    \vspace{-1.5em}
    \label{fig:system_overview}
\end{figure*}
}

\section{LLMs for Polymer Design}
\label{sec:llm_for_polymer}

Polymer design, a challenging task requiring experts, can be accelerated by LLMs but are limited by lack of model's grounding in polymer domain. Chemistry-aligned models like \Llamole{} \cite{llamaole}, \ChemLLM{} \cite{chemllm}, \etherzero{} \cite{ether0}, and \LlamaSmol{} \cite{llasmol} generalize to general chemistry tasks (retrosynthesis, property prediction etc.) but remain ineffective for polymers due to complexity and under-representation. Hence polymer-specific LLMs are needed.

\begin{figure*}[t!]
    \centering
    \includegraphics[width=0.8\textwidth]{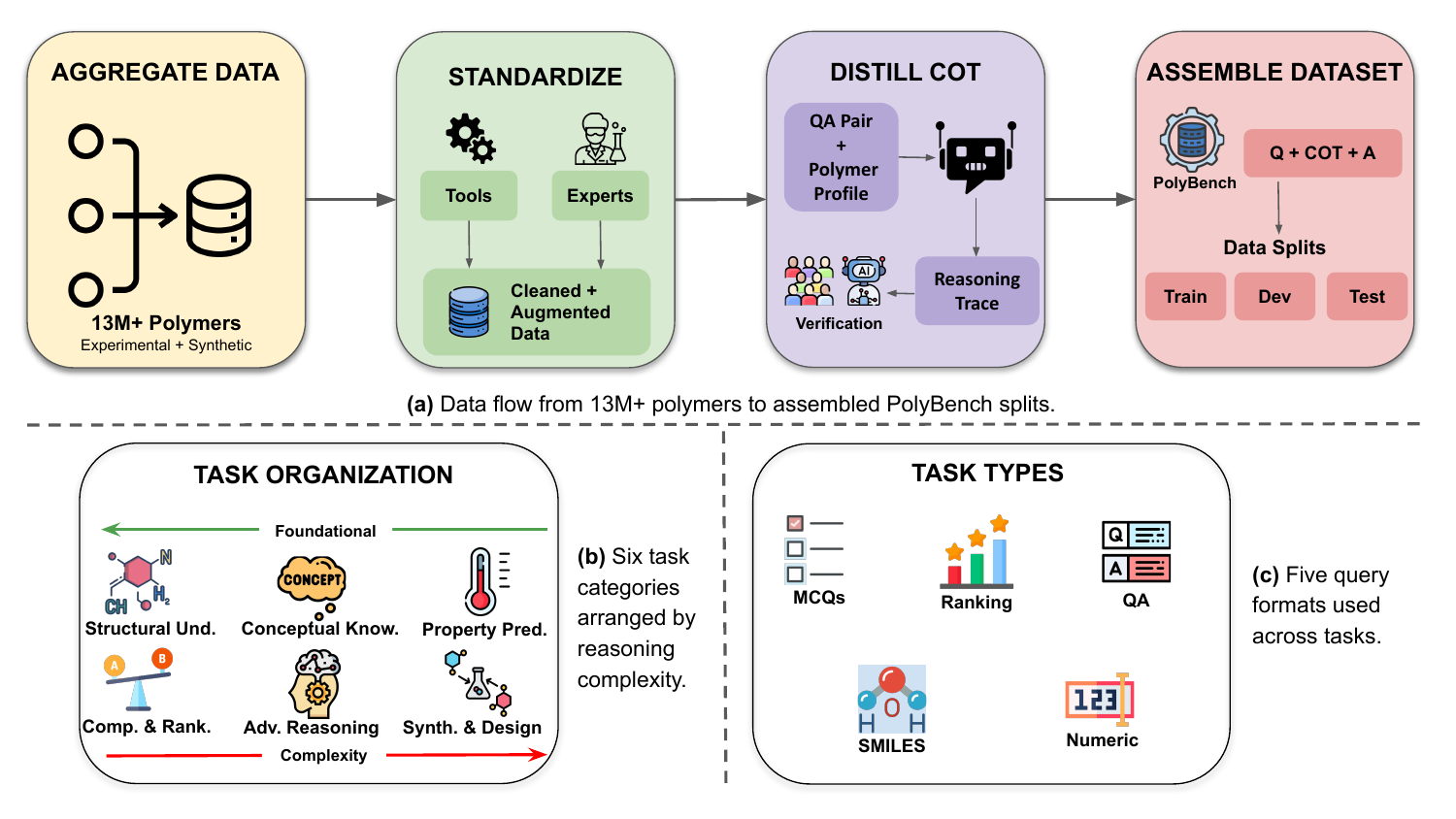}
    \caption{\textbf{PolyBench overview.} \textbf{(a)} \textit{Creation pipeline.} We aggregate data from open-source databases covering $13$M+ polymers, standardize notations and properties using \texttt{RDKit} and SME input, and distill chain-of-thought traces by prompting teacher models with knowledge-augmented polymer profiles. Reasoning traces samples are verified through automated judgment and manual review before being assembled into \texttt{(instruction, question, CoT, solution)} tuples with non-overlapping train/dev/test splits. \textbf{(b)} \textit{Task organization.} Six task categories arranged by reasoning complexity, from foundational skills to end-to-end design. \textbf{(c)} \textit{Task types.} Five query formats for tasks.}
    \vspace{-1.5em}
    \label{fig:system_overview}
\end{figure*}

\subsection{Challenges}
There are four main challenges in developing LLM-based solutions:\\
\noindent\textbf{1) Polymer Representation} Polymers are encoded using syntactic representations like \smiles{}, which capture their repeating and hierarchical structures. Reasoning over these representations must be precise as a slight change yields a different polymer. While LLMs handle complex syntax like programming languages well, we find that even chemistry-aligned models struggle with \smiles{}\footnote{Graph-based architectures (e.g., GNNs~\cite{queen_polymer_2023}) are better suited for \smiles{}~\cite{SMILES_paper}, but our goal is to improve LLMs' ability to \textit{natively} reason over these representations as a more scalable solution.}.
\noindent\textbf{2) Diverse Foundational Skills} Polymer design requires multiple skills and foundational knowledge - structural interpretation, property reasoning, comparison/ranking under constraints, and synthesis planning - rather than a single isolated capability. 
Teaching these skills to LLMs requires translating them to scalable, evaluatable task categories, which is challenging as the corresponding data remains siloed across multiple sources.

\noindent\textbf{3) Task Complexity} Polymer design is complex and multi-objective: design-related needs require reasoning across multiple properties, constraints, and goals (e.g., property trade-offs plus synthesis feasibility). This also demands \textit{cross-scale} reasoning - connecting local substructures (monomers / functional groups) and global factors (chain length, topology, architecture) to design task needs. 

\noindent\textbf{4) Compositional Reasoning \& Generalization} 
To effectively navigate this complexity, models require compositional multi-step capabilities -- for example generating candidate polymer modifications might involve understanding the given polymer structure, propose edits, make comparative property analysis and produce synthesis reasoning. Training and evaluation must therefore span these interconnected task types and test generalization to new polymers, new properties, and their conjunctions. We also need explicit \emph{reasoning traces} that provide necessary intermediate steps that guide model towards correct designs.

\noindent These challenges show that LLMs for polymer design require: (i) comprehensive multi-task training data spanning structural understanding, property inference, and synthesis planning, and (ii) explicit chain-of-thought reasoning mirroring how scientists decompose and solve complex problems. To this end, we introduce \dataset{}, a comprehensive benchmark suite for polymer alignment and evaluation.
\eat{\begin{table}[t]
\centering
\small
\setlength{\tabcolsep}{4pt}
\renewcommand{\arraystretch}{1.05}
\begin{tabularx}{\columnwidth}{l r l}
\toprule
\textbf{Source} & \textbf{\# Instances} & \textbf{Polymer/Monomer} \\
\midrule
PolymersML        & 18{,}667     & Homopolymers \\
Bicerano Polymers & 315          & Polymers \\
MTL Khazana       & 3{,}655      & Polymers \\
PolymerDB         & 224          & Polymers \\
Chembl            & 2{,}851      & Monomers \\
ToPoRg-1342       & 1{,}342      & Polymers \\
PN2S              & 11{,}165     & Co/Homo-Polymers \\
OMG Monomers      & 77{,}281     & Monomers \\
OMG Polymers      & 12{,}886{,}131 & Polymers \\
\bottomrule
\end{tabularx}
\caption{\label{tab:polymer-sources}
Polymer data sources and coverage.
}
\vspace{-6pt}
\end{table}}

\section{\dataset{}}
\label{sec:polybench}

We build \dataset{}, a large-scale, broad-coverage dataset to train and test LLMs for polymer design. We have three desiderata for \dataset{}: \textit{First}, since polymer design is a workflow rather than a single task -- we want the testbed to cover a range of polymer tasks with varied complexities - from foundational skills (structural interpretation, conceptual knowledge, property reasoning) to downstream design-oriented queries (e.g., comparison/ranking and synthesis feasibility). \textit{Second}, because polymer design is a highly specialized domain where "correct" answers must adhere to specific syntax and design constraints, we require the generation of explicit reasoning traces to make the basis for answers inspectable and enable failure diagnosis - the ability to trace back through the decision chain and identify types of failure. 
\textit{Third}, because polymers span across diverse classes/families (e.g., polyesters, polythene \cite{poly_classes_1, poly_classes_2, poly_classes_3}), our evaluation needs to reflect generalization to these new unseen polymers and tasks. Manually curation of such data is expensive. 


How do we create such high-quality broad-coverage data? Our main idea is to first combine open-source and experimentally verified polymer data on fundamental descriptions, properties and synthesis routes (Appendix~\ref{app:dataset-integration}) with large coverage. Then, using instruction following LLMs, we obtain reasoning trajectories grounded in real-world data (Subsection~\ref{subsec:reasoning-distillation}). For coverage across different generalization capabilities, we organize \dataset{} into tasks with test splits yielding valuable diagnostics (Subsection~\ref{subsec:task-organization}).

\subsection{Polymer Tasks Design and Organization}
\label{subsec:task-organization}
We derive our task categories from subject-matter expert (SME) input and common day-to-day workflows in polymer research, where scientists iteratively interpret structures, apply domain concepts, screen candidates by properties, compare and filter under constraints, and plan feasible synthesis routes. Using this workflow, we identify and organize six \textit{reasoning} task categories of \dataset{} - spanning across foundational skills to end-to-end design - that together capture the core capabilities a model needs to support polymer design-related tasks. These are: (i) \textbf{\textit{Structural Understanding}} (polymer representations, syntax, and structure parsing), (ii) \textbf{\textit{Conceptual Knowledge}} (core polymer science principles), (iii) \textbf{\textit{Property Prediction}} (structure$\rightarrow$property estimation), (iv) \textbf{\textit{Property Comparison \& Ranking}} (multi-criteria screening under constraints), (v) \textbf{\textit{Advanced Property Reasoning}} (mechanistic and trade-off reasoning), and (vi) \textbf{\textit{Synthesis \& Design}} (monomer/reaction selection and constrained design). We provide additional details and examples in Appendix~\ref{app:task_design_details}.

\subsection{Task Generation}
\label{subsec:task-generation}

We generate tasks from established polymer property databases to ensure broad coverage of polymers while maintaining scientific rigor. For each polymer, we construct a standardized \emph{polymer profile} by aggregating its structural representation (e.g., IUPAC/SMILES\footnote{In chemistry, IUPAC names are systematic, universally recognized names for chemical compounds, while SMILES are standardized text strings that encode molecular structure (atoms, bonds, and connectivity).}), experimentally reported properties, and computed structural attributes.

To support our task categories - we create tasks using two complementary processes: First, \emph{programmatic} generation converts profile entities directly into natural language QA - with constrained answer formats, covering structural queries, representation conversions, property targets, and structured synthesis decisions, as well as candidate screening. Second, \emph{distilled} generation prompts \claudeSonnet{} with the polymer profile(s) \emph{and} a category-conditioned instruction and examples to produce QA pairs grounded in the same underlying data in more open-ended formats. We use \claudeSonnet{}  based on preliminary prompting experiments and manual evaluation, where it consistently outperformed other LLMs at task generation.

Each category uses one or multiple task formats—open-ended questions, ranking, multiple-choice, numerical predictions, and SMILES generation. Within each category, tasks vary in complexity from single-step reasoning problems to multi-step problems requiring reasoning over different threads of information in a single query. Task-wise question types are shown in Table~\ref{tab:task_formats}. Tasks are generated after applying train/dev/test split to prevent polymer overlap across splits and support out-of-distribution evaluation. 
\dataset{}'s train set has more than $125$K questions spanning the six tasks, with dev and test sets of over $1.5$K and $2.3$K questions respectively (details in Appendix~\ref{app:dataset_stats}). Additionally, to support our compositionality analysis, we generate sub-questions, which serve as our diagnostic probes to identify model failures. Details are listed in Appendix~\ref{app:additional_diagnostics}.

\subsection{Knowledge-Augmented Distillation}
\label{subsec:reasoning-distillation}


Our task instances provide instructions and answers, but training models for polymer design also requires reasoning traces~\cite{wei2022chain} that reflect how polymer chemists justify decisions for unfamiliar polymers. Directly distilling CoT~\cite{hsieh-etal-2023-distilling, magister-etal-2023-teaching} from frontier LLMs is unreliable, since these models often lack polymer-specific knowledge; moreover, teacher limitations can transfer to the student. We mitigate this by grounding the teacher on exhaustive polymer profiles and gold targets, so it primarily explains provided evidence rather than generating facts from memory, reducing hallucination risk in the distilled traces. Our \emph{knowledge-augmented} distillation pipeline consists of: (1) \textbf{Knowledge injection} -- we augment each prompt with the relevant polymer profile (source-dataset fields and RDKit-derived structural attributes) so reasoning is grounded in complete structural and property evidence; (2) \textbf{Reasoning structure} -- we provide high-level guidance to decompose the task and produce a structured justification before the final answer (Appendix~\ref{app:cot_prompts}); and (3) \textbf{Automated verification} -- we run fact-checking with \gptfive{} on a subset of data from our validation set -- this verification achieves $\sim80\%$ accuracy in identifying correct reasoning chains. Manual validation suggests most errors are minor (e.g., incomplete explanations, imprecise terminology, or missing intermediate steps) rather than severe hallucinations. Importantly, these issues are confined to the generated CoT traces -- final answers come from gold labels (or teacher-generated answers for open-ended tasks). We therefore retain all traces with verification labels rather than filtering, which is computationally feasible at our dataset scale ($125$K+ instances). We further discuss the quality of distilled CoT in Section~\ref{sec:qual_of_cot}, including automated and manual verification, and types of CoT errors. We use \gptFourO{} and \claudeSonnet{} as teacher models to generate high-quality reasoning traces for \dataset{} using our knowledge-augmented distillation pipeline; full prompts are in Appendix~\ref{app:cot_prompts}.
\section{Experimental Setup}
\label{exp_setup}
We use \dataset{} as a polymer benchmark to assess the performance of existing open-source and frontier LLMs and compare them against models that are fine-tuned on \dataset{} train and dev sets and external relevant benchmarks. 

\subsection{Models and Setup}
\label{pb_evalmodels}

Using \dataset{} train and dev sets, we finetune \qwen{}, \qwenbig{}, \qwenbigger{}~\cite{qwen25model}, and \microsoftphi{}~\cite{phi4model}. We also finetune \ChemLLM{}~\cite{chemllm} with \dataset{} to assess the domain adaptability of chemistry domain models to polymer design tasks. We train each model using LoRA~\cite{hu2022lora}, with $r$=$16$ and $\alpha$=$32$, for $3$ epochs with varying batch sizes and learning rates per model (see Appendix~\ref{app:hyperparams}). For baselines, we use off-the-shelf versions of finetuned models and the following open-source chemistry domain-aligned models: \Llamole{}~\cite{llamaole}, \LlamaSmol{}~\cite{llasmol}, \etherzero{}~\cite{ether0}, \ChemLLM{}~\cite{chemllm} (more details in Appendix~\ref{app:domain_aligned_baselines}). We also compare against frontier models: \gptFourO{}~\cite{openai2024gpt4o}, \gptfivefour{}~\cite{singh2025openaigpt5card} and \claudeSonnetfoursix{}~\cite{anthropic2026claude46sonnet}\footnote{Knowledge cutoffs: \gptFourO{}: October 2023, \gptfivefour{}: August 2025, \claudeSonnetfoursix{}: August 2025}.

We do not include polymer-specific LLM baselines because prior polymer-focused models are discriminative models and optimized for narrow, task-specific interfaces (e.g., property regression), making them incompatible under our unified multi-task, instruction-following setting with natural-language outputs and reasoning traces. 

\subsection{Benchmark and Metrics}
We evaluate on two testbeds. First is held-out \dataset{} test set, containing unseen polymers, properties, and task variations. Second are the following external testbeds:  \texttt{Block Polymers}~\cite{jablonkagroup_block_polymers_morphology} contains questions about block co-polymers; \texttt{Llamole}~\cite{llamaole} contains molecule synthesis tasks; \texttt{ChemData}, which has polymer-related questions; \texttt{PolyReal} contains polymer experimentation questions. More details in Appendix~\ref{app:external_benchmarks}. Questions with overlapping polymers with \dataset{} are removed and we evaluate our models with a total of $1780$ testcases.

Tasks have mixed formats (e.g. QA, MCQ), shown in Table ~\ref{tab:task_formats}. For \textit{open-ended question answering}, we report \texttt{ROUGE-L(RgL)}~\cite{lin-2004-rouge} and \texttt{LLM-as-a-Judge(Jdg)}~\cite{zheng2023judging} scores (1--7 Likert) using \gptOSS{}~\cite{openai2025gptoss120bgptoss20bmodel}; for \textit{numeric} tasks, we report \texttt{Pearson Correlation Coefficient (r)} and \texttt{Mean Absolute Error (MAE)}; for \textit{counting} tasks, we report \texttt{Exact Match (EM)}; we report accuracy for multi-choice questions (\texttt{MCQ}) and Kendall's Tau Correlation (pairwise) for ranking tasks (\texttt{Pair}). For \smiles{} generation, we report \texttt{Similarity(Sim.)}~\cite{tanimoto_similarity_smiles} and \texttt{Validity(Val.)} using RDKit metrics to assess structural correctness and chemical validity of generated \smiles{}; we additionally report \texttt{Synthetic Accessibility(SA) scores}~\cite{sa-scores} using RDKit in Appendix~\ref{app:smiles_findings}. More details in Appendix~\ref{app:eval_metrics_detailed}. 

\section{Main Results}
\begin{table*}[!htbp]
\centering
\footnotesize
\begin{tabular}{lccccccccccccc}
\toprule
 & \multicolumn{3}{c}{\textbf{Structural}} & \multicolumn{2}{c}{\textbf{Polymer}} & \multicolumn{2}{c}{\textbf{Property}} & \multicolumn{2}{c}{\textbf{Prop. Comp.}} & \multicolumn{2}{c}{\textbf{Adv. Prop.}} & \multicolumn{2}{c}{\textbf{Design \&}} \\
 
 & \multicolumn{3}{c}{\textbf{Understanding}} & \multicolumn{2}{c}{\textbf{Concepts}} & \multicolumn{2}{c}{\textbf{Prediction}} & \multicolumn{2}{c}{\textbf{\& Ranking}} & \multicolumn{2}{c}{\textbf{Reasoning}} & \multicolumn{2}{c}{\textbf{Synthesis}} \\
 
 \cmidrule(lr){2-4} \cmidrule(lr){5-6} \cmidrule(lr){7-8} \cmidrule(lr){9-10} \cmidrule(lr){11-12} \cmidrule(lr){13-14}
\textbf{Model} & \textbf{EM} & \textbf{RgL} & \textbf{Jdg} & \textbf{RgL} & \textbf{Jdg} & \textbf{r} & \textbf{MAE ($\downarrow$)}  & \textbf{MCQ} & \textbf{Pair} & \textbf{RgL} & \textbf{Jdg} & \textbf{Sim.} & \textbf{Val.} \\
\midrule
\multicolumn{14}{l}{\textit{Off-the-shelf}} \\
\qwen{} & 0.47 & 0.12 & 3.78 & 0.23 & 4.83 & 0.78 & 73.60 & 0.43 & 0.58 & 0.27 & 4.04 & 0.17 & 0.68 \\
\qwenbig{} & 0.50 & 0.11 & 4.15 & 0.23 & 5.00 & 0.93 & 48.80 & 0.55 & 0.69 & 0.25 & 4.56 & 0.18 & 0.81 \\
\microsoftphi{} & 0.67 & 0.12 & 4.43 & 0.23 & 5.67 & 0.91 & 46.20 & 0.61 & 0.70 & 0.28 & 5.19 & 0.23 & 0.65  \\
\qwenbigger{} & 0.71 & 0.13 & 3.33 & 0.22 & 5.23 & 0.46 & 32.21 & 0.65 & 0.73 & 0.14 & 4.36 & 0.31 & 0.81 \\
\midrule
\multicolumn{14}{l}{\textit{Domain aligned}} \\
\Llamole{} & 0.21 & 0.09 & 2.50 & 0.18 & 2.80 & 0.13* & inf & 0.39 & 0.56 & 0.22 & 2.90 & 0.24 & 0.62  \\
\LlamaSmol{} & 0.27 & 0.08 & 3.34 & 0.17 & 3.68 & 0.07* & 181.70 & 0.20 & 0.49 & 0.24 & 3.56 & 0.16 & 0.31  \\
\etherzero{} & 0.38 & 0.07 & 3.26 & 0.15 & 3.13 & 0.10* & inf & 0.46 & 0.51 & 0.15 & 3.19 & 0.18 & 0.52  \\
\ChemLLM{} & 0.47 & 0.11 & 3.34 & 0.22 & 5.00 & 0.76 & 78.00 & 0.42 & 0.47 & 0.24 & 3.56 & 0.20 & 0.25  \\
\midrule
\multicolumn{14}{l}{\textit{Closed-source}} \\
\gptFourO{} & 0.60 & 0.29 & 3.71 & 0.27 & 5.00 & 0.98 & 22.70 & 0.69 & 0.86 & 0.33 & 4.73 & 0.16 & 0.11   \\
\claudeSonnetfoursix{} & 0.47 & 0.26 & 4.74 & 0.27 & 5.80 & 0.98 & 20.50 & 0.66 & 0.85 & 0.30 & 4.88 & 0.26 & 0.29 \\
\gptfivefour{} & 0.60 & 0.10 & \textbf{5.10} & 0.17 & 6.38 & 0.92 & 35.05 & \textbf{0.70} & \textbf{0.87} & 0.18 & 5.68 & 0.28 & 0.58  \\
\midrule
\multicolumn{14}{l}{\textit{\dataset{}}} \\

\qwen{} & 0.53 & \cellcolor{lime}0.27 & \cellcolor{lime}4.53 & \cellcolor{lime}0.24 & \cellcolor{lime}5.00 & \cellcolor{lime}0.93 & \cellcolor{lime}39.50 & 0.49 & 0.60 & 0.25 & \cellcolor{lime}4.96 & \cellcolor{green}0.64 & 0.72 \\

\quad\texttt{+ CoT} & 0.60 & \cellcolor{lime}0.24 & 3.91 & \cellcolor{green}0.29 & \cellcolor{lime}4.50 & \cellcolor{lime}0.91 & 47.10 & 0.48 & 0.57 & 0.26 & 4.67 & \cellcolor{green}0.43 & 0.62  \\

\ChemLLM{} & 0.47 & \cellcolor{lime}0.13 & 4.10 & \cellcolor{lime}0.27 & \cellcolor{lime}5.50 & \cellcolor{green}0.98 & \cellcolor{lime}19.90 & 0.59 & 0.69 & \cellcolor{green}0.35 & 4.61 & \cellcolor{green}0.74 & \cellcolor{green}\textbf{0.95} \\

\quad\texttt{+ CoT} & \cellcolor{lime}0.67 & \cellcolor{lime}0.32 & \cellcolor{lime}4.30 & \cellcolor{green}0.33 & \cellcolor{lime}6.25 & \cellcolor{green}0.98 & \cellcolor{lime}20.30 & 0.54 & 0.50 & \cellcolor{green}0.35 & \cellcolor{lime}5.18 & \cellcolor{green}0.63 & 0.66  \\

\qwenbig{} & \cellcolor{lime}0.70 & 0.11 & 3.46 & \cellcolor{lime}0.27 & \cellcolor{lime}6.33 & \cellcolor{lime}0.97 & \cellcolor{lime}37.50 & 0.50 & \cellcolor{lime}\textbf{0.87} & \cellcolor{lime}0.31 & \cellcolor{lime}4.86 & \cellcolor{green}0.68 & \cellcolor{green}0.81 \\

\quad\texttt{+ CoT} & \cellcolor{green}\textbf{0.87} & \cellcolor{lime}0.37 & \cellcolor{lime}4.63 & \cellcolor{green}\textbf{0.31} & \cellcolor{green}\textbf{6.52} & \cellcolor{green}0.98 & \cellcolor{lime}21.70 & \cellcolor{lime}0.63 & 
\cellcolor{lime}0.73 & \cellcolor{green}0.39 & \cellcolor{lime}5.10 & \cellcolor{green}0.70 & \cellcolor{green}0.94  \\

\microsoftphi{} & \cellcolor{green}0.80 & \cellcolor{lime}0.13 & \cellcolor{lime}4.43 & \cellcolor{green}0.29 & \cellcolor{lime}6.00 & \cellcolor{green}0.98 & \cellcolor{lime}\textbf{16.80} & 0.57 & \cellcolor{lime}0.85 & \cellcolor{green}0.33 & \cellcolor{lime}5.08 & \cellcolor{green}\textbf{0.82} & 0.66  \\

\quad\texttt{+ CoT} & \cellcolor{green}0.80 & \cellcolor{lime}0.18 & \cellcolor{lime}4.72 & \cellcolor{green}\textbf{0.31} & \cellcolor{lime}6.33 & \cellcolor{green}\textbf{0.98} & \cellcolor{lime}18.10 & \cellcolor{lime}0.65 & \cellcolor{lime}0.71 & \cellcolor{green}0.38 & \cellcolor{lime}5.21 & \cellcolor{green}0.75 & \cellcolor{green}0.82 \\

\qwenbigger{} & \cellcolor{green}0.79 & \cellcolor{green}\textbf{0.32} & 3.36 & \cellcolor{lime}0.26 & 4.87 & 0.54 & 80.05 & \cellcolor{lime}0.65 & \cellcolor{lime}0.73 & \cellcolor{lime}0.28 & 4.36 & \cellcolor{lime}0.31 & 0.78 \\
\quad\texttt{+ CoT} & \cellcolor{green}0.84 & \cellcolor{lime}0.15 & \cellcolor{lime}4.66 & \cellcolor{green}0.28 & \cellcolor{lime}5.77 & \cellcolor{lime}0.96 & 26.07 & 0.55 & \cellcolor{lime}0.74 & \cellcolor{green}\textbf{0.46} & \cellcolor{green}\textbf{6.38} & \cellcolor{green}0.64 & \cellcolor{green}0.92  \\

\bottomrule
\end{tabular}

\caption{
Model performance metrics on \dataset{}, classified by task.
\textcolor{lime}{\rule{1.5ex}{1.5ex}} denotes cases where the corresponding model surpasses open-source models (off-the-shelf and domain-aligned), whereas \textcolor{green}{\rule{1.5ex}{1.5ex}} indicates where \dataset{} models outperform all model categories. The best performance is \textbf{bolded}. 
}
\label{tab:main_results}
\vspace{-2.3em}
\end{table*}
Table \ref{tab:main_results} shows the performance on test set of \dataset{}, divided across each reasoning task type\footnote{inf: MAE undefined. *: Pearson r not statistically significant (p >= 0.05)} in Section~\ref{subsec:task-organization}. We observe the following.

\subsection{\dataset{} is challenging for LLMs} 
Results in Table \ref{tab:main_results} highlight multiple challenges for current LLMs on polymer design task. 


\textbf{Off-the-shelf models fare poorly across all tasks.} Base models struggle across the board, especially when the task requires \smiles{} representations. This suggests a lack of exposure to polymer notation and design processes during pretraining of these models. 
\textbf{General chemistry alignment is inadequate for \dataset{}.} Chemistry-aligned models (\Llamole{}, \LlamaSmol{}, \etherzero{}, \& \ChemLLM{}) under-perform on almost all tasks in \dataset{}, suggesting that general chemistry knowledge (where mostly small molecules are focused) does not directly transfer to polymers. This is further supported by \ChemLLM{} trained on \dataset{}, which consistently outperforms its base version and shows greater domain adaptation - indicating that polymer-specific supervision is needed to close this gap.
\textbf{Design tasks are challenging even for frontier LLMs.} Frontier models perform strongly on several foundational tasks, but struggle on polymer design tasks, which require generating chemically valid \smiles{} for polymers while satisfying property, structure, and synthesis requirements. To further evaluate generated \smiles{}, we add synthetic accessibility(SA) scores~\cite{sa-scores}. We observe that while some frontier models generate molecules with higher SA scores, \dataset{}-trained models achieve higher similarity to target polymers while maintaining much lower SA scores (Appendix~\ref{app:smiles_findings}).    
Together these show the large room for improvement in the reasoning and compositional skill integration - especially when models must connect structured representations to properties and constraints and produce valid, consistent output.
\vspace{-0.5em}
\subsection{\dataset{} Training Substantially Improves Performance}
One ablation involves training all our base- and one domain-aligned model using \dataset{} train set. We observe that \dataset{} models achieve the best performance across all tasks compared to similar-sized baselines (off-the-shelf and domain-aligned) with the bigger models (\texttt{14B \& 32B} variants), showing the best gains.

Table \ref{tab:main_results} also shows that the effect of Chain-of-Thought training (\texttt{+CoT} rows) scales with model size. 
For the smaller models \qwen{} and \ChemLLM{}, \texttt{CoT} improves performance on $3$ and $6$ of $13$ metrics respectively, showing limited gains~\cite{Li2025SmallMS}. 
However, for the larger \qwenbig{}, \microsoftphi{} and \qwenbigger{}, we see that our \texttt{CoT} training leads to better performance on $8$ to $12$ of $13$ metrics. Interestingly, we see that \texttt{CoT} benefits some tasks over others. Clearest gains appear in compositional tasks (such as \taskFivereasoning{}) that require reasoning over multiple properties. However, improvements are limited in tasks requiring \emph{precise} quantitative targets (e.g., \taskThreeproppred{}), where success depends on accurately mapping structure to a numeric value.
Our intuition is that reasoning for such tasks is more complex. Therefore, \texttt{CoT} fails to capture the reasoning leading to correct answer and just adds verbosity. This aligns with prior evidence in NLP literature~\cite{dziri2023faith, sprague2025to}: training improves performance on multi-step compositional reasoning tasks, while offering limited—or even negative—gains for tasks that can be solved through direct pattern matching. 

\begin{table}[!tbp]
\footnotesize
\centering
\setlength{\tabcolsep}{5pt}
\begin{tabular}{lccccc}
\toprule
\textbf{Model} & \textbf{Blk} & \textbf{ChemD} & \textbf{Llml} & \textbf{PolyR} & \textbf{Avg} \\
\midrule
\multicolumn{6}{l}{\textit{Off-the-shelf}} \\
\microsoftphi{}      & 1.74 & 5.96 & 1.36 & 4.44 & 3.37 \\
\qwenbig{}     & 1.93 & 5.83 & 1.48 & 4.31 & 3.39 \\
\qwenbigger{}     & 2.44 & 6.10 & 1.49 & 4.55 & 3.62 \\
\midrule
\multicolumn{6}{l}{\textit{Domain aligned}} \\
\ChemLLM{}   & 2.35 & 5.94 & 1.44 & 3.85 & 3.40 \\
\LlamaSmol{} & 1.56 & 5.37 & 2.16 & 1.50 & 2.60 \\
\etherzero{} & 1.63 & 1.37 & 1.14 & 1.16 & 1.27 \\
\midrule
\multicolumn{6}{l}{\textit{Closed-source}} \\
\gptFourO{}  & 1.89 & 6.05 & 1.64 & 4.93 & 3.63 \\
\gptfivefour{}   & 3.03 & \textbf{6.61} & 2.29 & \textbf{5.95} & \textbf{4.47} \\
\claudeSonnetfoursix{}    & 2.66 & 6.20 & 3.03 & 4.09 & 3.99 \\

\midrule
\multicolumn{6}{l}{\textit{PolyLM}} \\
\microsoftphi{}      & \cellcolor{green}3.07 & 5.96 & 1.92 & \cellcolor{lime}4.46 & \cellcolor{lime}3.85 \\
\quad \texttt{+ CoT} & \cellcolor{green}\textbf{3.14} & 6.03 & \cellcolor{green}2.43 & \cellcolor{lime}4.63 & \cellcolor{lime}4.06 \\
\qwenbig{}     & \cellcolor{green}2.69 & \cellcolor{lime}6.11 & 2.11 & 4.47 & \cellcolor{lime}3.84 \\
\quad \texttt{+ CoT} & 2.04 & 5.97 & 2.16 & \cellcolor{lime}4.56 & \cellcolor{lime}3.68 \\
\qwenbigger{}      & \cellcolor{green}3.07 & \cellcolor{lime}6.11 & 2.11 & \cellcolor{lime}4.61 & \cellcolor{lime}3.97 \\
\quad \texttt{+ CoT} & \cellcolor{green}3.06 & \cellcolor{lime}6.16 & \cellcolor{green}\textbf{2.50} & \cellcolor{lime}4.83 & \cellcolor{lime}4.14 \\
\bottomrule
\end{tabular}





\caption{Performance on external polymer benchmarks (Likert 1–7 scale). For highlighted \dataset{} model values,
\textcolor{lime}{\rule{1.5ex}{1.5ex}} denotes where the corresponding surpasses open-source models(off-the-shelf and domain-aligned) whereas \textcolor{green}{\rule{1.5ex}{1.5ex}} denotes where \dataset{} models beat all other category models. The best performance is \textbf{bolded}.}
\label{tab:external_results_with_pass5}
\vspace{-2em}
\end{table}

\subsection{\dataset{} Training helps on External Datasets}
Table~\ref{tab:external_results_with_pass5} evaluates a subset of \dataset{} trained models on three external polymer benchmarks for a stronger out-of-distribution setting, using LLM-as-a-Judge (Likert 1--7). \dataset{}-models outperform off-the-shelf and domain-aligned baselines across all benchmarks, and remain competitive with closed-source models overall: they lead on Blk and are close on Llml, while ChemD (mostly MCQ and single word/sentence question answering), and PolyR (broad scientific knowledge about real-world polymer applications) - remains more favorable to frontier models. 
\vspace{-0.5em}
\section{Quantifying the Compositionality Gap} 
\label{sec:error_analysis} 
\vspace{-0.5em}
Standard metrics fail to explain \textit{why} models fail on polymer design tasks. Hence, using our diagnostic set from \ref{subsec:task-organization}, we run three core experiments using \dataset{}-trained \microsoftphi{}\texttt{+CoT}: (i) \textbf{Exp~1 (Skill Gap)} tests whether the model can answer atomic sub-questions in isolation; (ii) \textbf{Exp~2 (Skill Selection \& Accuracy)} measures whether the model's \texttt{CoT} covers the required sub-questions (recall) and whether the included steps are correct (precision); and (iii) \textbf{Exp~3 (Compositionality Gap)} evaluates end-task performance when the model is given additional context (sub-questions (SubQs) or, sub-questions and gold answers (SubQs + gold)). Our findings suggest that models often know the necessary pieces (high SubQ precision in Exp~1), but struggle to select and propagate the right constraints/steps in their own reasoning (low recall in Exp~2), and even when given SubQ/SubQA as context (Exp~3), composing everything into a consistently constraint-satisfying solution remains brittle -- evidence of a compositionality gap rather than pure knowledge missing. Additional details are in Appendix \ref{app:additional_diagnostics}.

\section{Validity of LLM-as-a-Judge} 
\label{sec:validity_of_llm_as_judge}
\vspace{-0.5em}


We use \texttt{LLM-as-a-Judge} to evaluate open-ended QA questions (3 out of 6 tasks from \dataset{}, and external benchmarks). We additionally report ROUGE-L, and our findings show that trends across the two metrics consistently agree. Further, our \texttt{LLM-as-a-Judge} vs \texttt{Human} scores calibrations shows a strong correlation between the two scores. We also probe our \texttt{LLM-as-a-Judge} for length and stylistic bias for length and stylistic bias by comparing models with distinct output styles - we see a non-monotonic length-score relationship (Table~\ref{tab:judge-length-by-score}): verbose responses do not translate into higher scores. More details in Appendix~\ref{app:additional_diagnostics}.
\vspace{-0.5em}
\section{Quality of Distilled CoT}
\label{sec:qual_of_cot}
\vspace{-0.5em}
Using a stratified sample of 300 examples from our training and validation set, we conduct an automatic evaluation of generated CoT traces, followed by a manual error analysis to characterize the error. Our findings reveal 80.67\% as fully correct (score 5 on a 1--5 Likert scale), with the remaining 19.33\% showing dominant errors (Table~\ref{tab:cot-errors}) are missing steps and incomplete reasoning (Details in \ref{app:additional_diagnostics}). 

\section{Conclusion} 
\vspace{-0.5em}
Polymer design requires deep understanding of polymer structures and properties, something that general-purpose and chemistry-domain LLMs lack. 
Addressing this limitation, we develop \dataset{} -- a comprehensive benchmark for polymer understanding spanning six important task types from structure parsing to synthesis, with CoT traces grounded in hard experimental data. Training and testing a suite of 7B-14B models on \dataset{} highlights the effectiveness of \dataset{} as a training dataset, where trained models outperform baselines and remain competitive with frontier LLMs. We believe this work advances polymer informatics by providing a benchmark to the community and offers a blueprint for building science-domain LLMs.

\section*{Limitations}
We propose \dataset{}, a benchmark for training AI models for polymer domain. Using \dataset{}, we finetune models and test them against baselines on our own benchmark as well other established benchmarks. Despite our comprehensive evaluation pipeline, we are aware of the limitations of our work. 

Firstly, our dataset is augmented with LLM generated CoT trajectories. Even though these trajectories are generated with a lot of target polymer context, model hallucinations might seep in, resulting in noisy lower quality CoT. In some cases, the CoT might be too complex due to the additional context to make it a useful signal for downstream alignment of smaller LLMs. While the CoTs have been verified on stratified samples using both automated self-verification and manual review, there is still room for error. Hence, training on such trajectories can propagate the teacher LLM's misinterpretations and hallucinations into the aligned model. We note that reasoning traces may reflect teacher-model conventions and may over-emphasize certain explanation styles; we therefore include verification labels and discuss risks of imprinting/bias in this section.

Our \dataset{} currently builds upon the open-source datasets about polymers and their properties. The current set of tasks and queries are actually extensions building upon the tasks proposed by other researchers along with new tasks resulting from augmentations using tools like \texttt{RDKit}. While our tasks might be larger than previously proposed works which usually focus on single tasks or properties, our set of tasks is still limited by what is at hand. For example, data such as lab synthesis planning and action are not covered. \dataset{} does not also teach models how to use coding and/or polymer based coding frameworks in order to solve problems. Given the agentic framework that is popular nowadays, lack of tasks for teaching tool usage in polymer domain is also a significant limitation in terms of tasks.

Third, \dataset{} contains information only in the text modality. Multiple works have shown that incorporating other modalities can significantly improve performance in chemistry domain tasks. For instance, including images of the molecule or graphical information along with textual information can significantly boost performance in a number of downstream tasks. Also, provision of other modalities enables researchers to align multi-modal models for polymer domain.

Finally, our evaluation using \dataset{} was only done for single-step inference. Neither \dataset{} nor our evaluation pipeline was designed for agentic approach. Multiple recent studies have shown that LLMs offer significant performance gains when acting as agents, being able to do multi-step reasoning and utilize relevant tools. Given how comprehensive \dataset{} is in terms of polymer design tasks, being able to further enhance its design by incorporating agentic approach through relevant tools and training trajectories, could have enhanced the benchmark.

\section{Data and Model Release}
\label{app:release}
\dataset{} will be released under CC BY-NC 4.0 for non-commercial research use at \url{https://github.com/StonyBrookNLP/PolyBench}. Fine-tuned model weights will be released on Hugging Face under their respective base model licenses (MIT for Phi-4, Apache 2.0 for Qwen-2.5). CoT traces were generated via the official \claudeSonnet{} and \gptFourO{} APIs for research in a specialized domain (polymer design). We will release distilled CoT traces consistent with each provider's terms of service; where terms restrict redistribution, we will release only our post-processed, verified reasoning chains rather than raw teacher outputs.
\bibliography{custom_modified, datasources, models}

\clearpage
\appendix

\section{Polymer Dataset Sources}
\label{app:polymer-dataset-sources}

Table~\ref{tab:polymer-sources-detailed} summarizes all data sources used in creating \dataset{}, including their size, polymer types, and the information they provide. 

\paragraph{Experimental Property Databases.} The foundation of our dataset comprises experimentally validated polymer-property relationships from four key sources. PolymersML \cite{polymersml,polyinfo} (18,667 homopolymer instances) provides the most comprehensive property coverage spanning physical properties (density, specific volume), thermal properties (glass transition temperature, crystallization temperature), and mechanical properties (tensile strength, compression modulus). Bicerano Polymers \cite{bicerano} (315 instances) contributes both experimental and calculated values for key properties including glass transition temperature, density at 300K, and thermal expansion coefficients. MTL Khazana \cite{huan_polymer_2016} (3,655 instances) focuses on quantum-mechanical and electronic properties such as bandgap, electron affinity, ionization energy, and dielectric constant. PolymerDB~\cite{polymerdb} (224 instances), originally collected from the Polymer properties database(source in the link), provides fundamental molecular descriptors including molecular weight, molar volume, solubility parameters, and glass transition temperatures with associated property ranges.

\paragraph{Structural and Topological Data.} To enable comprehensive structural understanding, we incorporate ToPoRg \cite{toporg} (1,342 polymer instances) with explicit topological annotations (cyclic, linear) and graph-based representations (nodes, edges). ChEMBL \cite{chembl} (2,851 monomer instances) provides monomer SMILES with base representations and CXSMILES extensions marking attachment sites \cite{cxsmiles}. These sources provide the structural foundation necessary for syntax understanding and molecular parsing tasks.

\paragraph{Synthesis and Reaction Data.} PN2S~\cite{pn2s} contributes 11,165 instances documenting monomer-to-polymer transformations for both copolymers and homopolymers, specifying reactants, products, and reaction types. We supplement this with the Organic Materials Generator (OMG)~\cite{omg_polymers} databases, which comprise 77,281 characterized monomers and 12,886,131 synthetically generated polymers. From this large synthetic collection, we sample 10,000 polymers through stratified sampling across the 17 reaction types covered in OMG-Polymers, ensuring broad coverage of reactants (monomers), product polymers, and reaction types. We limit our sample to 10K (rather than the full 12M) because OMG polymers are synthetically generated rather than lab-verified. OMG Monomers categorizes monomers by functional groups (e.g., di-amine, di-isocyanate, di-ol, lactam, cyclic ether) and provides synthetic accessibility scores; OMG Polymers documents systematic polymerization reactions linking reactants to products. Together, these sources bridge individual monomers and final polymer structures through both documented experimental pathways and systematic synthetic exploration.

\subsection{Data Integration and Standardization}
\label{app:integration-and-standardization}
Combining multiple sources raises several issues in terms of variations along representation, units of measurement, and structural aspects. 

\paragraph{Standardization.} We address these through a two-step approach. For molecular standardization, we use RDKit \cite{rdkit} to automatically canonicalize all SMILES strings, ensuring consistent representations by removing redundant stereochemistry, neutralizing charged species, and standardizing tautomeric forms. For property standardization, we consulted with polymer scientists to establish standard units across all datasets (e.g., temperatures in Kelvin, density in g/cm³, solubility parameter in MPa$^{1/2}$). We also identify and remove duplicate entries by comparing canonical SMILES strings and consolidate property values from multiple sources when the same polymer appears in different datasets.

\paragraph{Computational Structural Augmentation.} Working with polymer scientists, we identified a comprehensive set of molecular descriptors relevant for polymer understanding. All polymer and monomer structures are processed through RDKit \cite{rdkit} to automatically extract these features, including basic molecular properties (molecular weight, heavy atom count), structural characteristics (rotatable bonds, ring counts, complexity indices), hydrogen bonding features (donors, acceptors), and functional group identifiers (alcohols, amines, carbonyls, halogens, etc.). These features provide consistent structural representations across all data sources and enable models to learn from interpretable chemical descriptors.


\begin{table*}[t]
  \centering
  \small
  \setlength{\tabcolsep}{6pt} 
  \begin{tabularx}{\textwidth}{l r l l >{\raggedright\arraybackslash}X}
    \hline
    \textbf{Source} & \textbf{Num of Instances} & \textbf{Polymers/Monomers} & \textbf{Known/Generated} & \textbf{Comments} \\
    \hline
    PolymersML        & 18,667    & Homopolymers      & Known     & Polymer $\rightarrow$ Property \\
    Bicerano Polymers & 315      & Polymers          & Known     & Polymer $\rightarrow$ Property \\
    MTL Khazana       & 3,655     & Polymers          & Known     & Polymer $\rightarrow$ Property \\
    PolymerDB         & 224      & Polymers          & Known     & Polymer $\rightarrow$ Property \\
    ChEMBL            & 2,851     & Monomers          & Known     & Attachment Sites [*] \\
    ToPoRg       & 1,342     & Polymers          & Known     & Polymer $\rightarrow$ topology \\
    PN2S              & 11,165    & Co/Homo-Polymers  & Known     & Monomer $\rightarrow$ Polymer with type of reaction \\
    OMG Monomers      & 77,281    & Monomers          & Known     & Monomers $\rightarrow$ polymerisation + structural features (cyclic, etc.) \\
    OMG Polymers      & 12,886,131 & Polymers          & Generated & Monomers $\rightarrow$ Polymers \\
    
    \hline
  \end{tabularx}
  \caption{\label{tab:polymer-sources-detailed}
    Polymer data sources, coverage and notes. 
  }
\end{table*}

\section{Data \& Task Details}

\subsection{Generalization and Dataset Statistics}
\label{app:dataset_stats}

Table~\ref{tab:poly-splits} summarizes dataset splits across all six polymer reasoning tasks, detailing unique polymers and associated question counts for train, development, and test sets.

\begin{table}[t!]
\centering
\footnotesize
\setlength{\tabcolsep}{3pt}
\renewcommand{\arraystretch}{1.05}
\begin{tabularx}{\columnwidth}{
  >{\raggedright\arraybackslash}X
  >{\raggedleft\arraybackslash}p{0.9cm}
  >{\raggedleft\arraybackslash}p{0.8cm}
  >{\raggedleft\arraybackslash}p{0.9cm}
  >{\raggedleft\arraybackslash}p{0.8cm}
  >{\raggedleft\arraybackslash}p{0.9cm}
  >{\raggedleft\arraybackslash}p{0.8cm}
}
\toprule
\textbf{Task} &
\multicolumn{2}{c}{\textbf{Train}} &
\multicolumn{2}{c}{\textbf{Dev}} &
\multicolumn{2}{c}{\textbf{Test}} \\
& \textbf{P} & \textbf{Q} & \textbf{P} & \textbf{Q} & \textbf{P} & \textbf{Q} \\
\midrule
StrUnd.     & 12K  & 27.9K & 3K   & 401 & 4K   & 449 \\
ConKnow     & 8.5K & 17.1K & 2K   & 240 & 2.7K & 47 \\
PropPred    & 8.5K & 4.5K  & 2K   & 70  & 2.7K & 448 \\
PropCompR   & 8.5K & 19.2K & 2K   & 291 & 2.7K & 600 \\
AdvReason   & 8.5K & 25.1K & 2K   & 374 & 2.7K & 125 \\
SynDes      & 5K   & 31.5K & 1.2K & 449 & 1.8K & 638 \\
\bottomrule
\end{tabularx}
\caption{\label{tab:poly-splits}
\dataset{} tasks splits and statistics. 
\textbf{StrUnd.} = Structural Understanding,  
\textbf{ConKnow} = Conceptual Knowledge,  
\textbf{PropPred} = Property Prediction,  
\textbf{PropCompR} = Property Comparison \& Ranking,  
\textbf{AdvReason} = Advanced Property Reasoning,  
\textbf{SynDes} = Synthesis/Design.  
\emph{P} = unique polymers, \emph{Q} = questions/prompts.
}
\end{table}

\subsection{Dataset Integration and Standardization}
\label{app:dataset-integration}

We collect $\sim13$ million experimental and synthetic data across multiple polymer benchmarks. These include (i) four experimentally verified datasets of physical, thermal, and mechanical properties, (ii) two sources providing topological information with \smiles{} representations covering molecule parsing; and (iii) two synthesis data sources covering reactants, products, and reaction types (details in Appendix~\ref{app:polymer-dataset-sources}).

However, merging multiple datasets is difficult due to inconsistencies in polymer representations and differing units. To mitigate this, we use \texttt{RDKit} \cite{rdkit} to automatically standardize \smiles{} representations, convert to consistent units, and remove duplicates. We also standardize descriptors (e.g., molecular weight, atom counts), structural properties (e.g., ring counts), and functional group identifiers (e.g., alcohols), along with their structural representations with domain expertise and information from \texttt{RDKit}. Please see Appendix~\ref{app:integration-and-standardization} for more details.

\subsection{Task Details}
\label{app:task_design_details}

\textbf{Structural Understanding:} This category tests the model's ability to comprehend polymer molecular architecture through four task types: (1) \textit{Syntax Understanding} evaluates whether models can interpret polymer notation systems, (2) \textit{Structural Understanding} assesses comprehension of polymer topology and composition, (3) \textit{Structural Understanding with Counting} requires quantitative analysis of structural components, and (4) \textit{Syntax-to-Structure Translation} tests the ability to convert between different polymer representation formats. These tasks use a question-answering format.

\textbf{Conceptual Knowledge:} Tasks evaluating fundamental polymer science knowledge, including \textit{Conceptual Understanding} of polymer principles and \textit{Structural Understanding with Reasoning}, which requires explaining the relationship between structure and function. These tasks are evaluated in a question-answering format.

\textbf{Property Prediction:} Direct numerical prediction tasks where models must predict quantitative polymer properties (e.g., glass transition temperature, tensile strength, thermal conductivity) from molecular structure. Performance is measured using numerical regression metrics.

\textbf{Property Comparison and Ranking:} Multiple-choice tasks requiring comparative reasoning about polymer properties, including: (1) \textit{Property Ranking} where models order polymers by property values, (2) \textit{Property Comparison} between polymer pairs, and (3) \textit{Design Reasoning with Structural Constraints} spanning multiple sub-tasks: filter-then-optimize selection, logical composition of requirements, constrained ranking under multiple criteria, and triple intersection of property constraints.

\textbf{Advanced Property Reasoning:} Complex reasoning tasks in a question-answering format: (1) \textit{Property Decomposition with Mechanistic Understanding} requires models to explain the mechanistic origins of polymer properties, (2) \textit{Property Inference with Structural Decomposition} involves inferring properties by analyzing structural components, and (3) \textit{Multi-Property Reasoning with Tradeoffs} assesses understanding of competing property requirements in polymer design.

\textbf{Synthesis and Design:} Practical polymer synthesis tasks including: (1) \textit{Product Prediction} (predicting polymer structure from monomers and reactions), (2) \textit{Monomer Identification} (identifying required monomers for target polymers), (3) \textit{Second Monomer Identification} (completing monomer pairs for copolymerization), and (4) \textit{Reaction Identification} (determining appropriate polymerization reactions). The first three tasks require generating valid \smiles{} representations, while reaction identification uses a question-answering format. We also include open-ended \textit{Design Reasoning with Structural Constraints} tasks.

Task formats across categories (QA, numeric, MCQ, ranking, SMILES) are summarized in Table~\ref{tab:task_formats}.

\begin{table}[!htbp]
\small
\setlength{\tabcolsep}{3pt}  
\centering
\begin{tabular}{lcccccc}
\toprule
\textbf{Task} & \rotatebox{90}{\textbf{QA}} & \rotatebox{90}{\textbf{Numeric}} & \rotatebox{90}{\textbf{MCQ}} & \rotatebox{90}{\textbf{Ranking}} & \rotatebox{90}{\textbf{SMILES}} \\
\midrule
\taskOnestrunderstanding{} & \checkmark & \checkmark & - & - & - \\
\taskTwopolymerconcepts{} & \checkmark & - & - & - & - \\
\taskThreeproppred{} & - & \checkmark & - & - & - \\
\taskFourpropcomprank{} & - & - & \checkmark & \checkmark & - \\
\taskFivereasoning{} & \checkmark & - & - & - & - \\
\taskSixdesign{} & - & - & - & - & \checkmark \\
\bottomrule
\end{tabular}
\caption{Task format distribution across PolyBench task categories.}
\label{tab:task_formats}
\end{table}

\subsection{Distilled vs.\ Programmatic Training Data}
\label{app:distilled_breakdown}

Of the $\sim$125K training questions in \dataset{}, $\sim$27K (approximately 22\%) are generated via teacher-model distillation (\claudeSonnet{} and \gptFourO{}), with the remaining 78\% generated programmatically from our knowledge base. Distillation is used selectively for task categories that require open-ended natural-language reasoning: 100\% of \textit{Conceptual Knowledge} (17.1K questions), 30\% of \textit{Advanced Property Reasoning} ($\sim$7.5K of 25.1K), and 10\% of \textit{Structural Understanding} ($\sim$2.8K of 27.9K). The remaining three categories -- \textit{Property Prediction}, \textit{Property Comparison and Ranking}, and \textit{Synthesis and Design} -- are generated entirely programmatically using structured templates over our polymer knowledge base, with ground-truth answers derived from experimental measurements or programmatic computation. The development and test splits follow a similar distribution of distilled and programmatic questions, with distilled content capped at 10\% per category to minimize reliance on teacher-generated content in evaluation.

We note that while \claudeSonnet{} is used for both task generation and as a test-time baseline, generation is grounded in verified polymer profiles with category-constrained prompts and manually-curated few-shot examples, and the remaining 78\% of tasks are generated programmatically with gold answers. This limits stylistic or content leakage. Furthermore, \dataset{}-trained models show consistent gains on external benchmarks (Table~\ref{tab:external_results_with_pass5}) that \claudeSonnet{} had no role in generating, indicating that our improvements are not primarily driven by teacher-induced stylistic alignment.

\subsection{Task Examples}
\label{app:task_examples}

Table \ref{tab:polybench_examples} presents representative examples from each polymer reasoning task, illustrating the diversity of input–output formats and reasoning requirements.

\section{Experimental Setup Details}
\label{app:exp_setup}

\subsection{Domain Aligned Baselines}
\label{app:domain_aligned_baselines}
To compare the performance of \dataset{} models, we select a set of weak and strong baseline models. The weak baseline models are the off-the-shelf unaligned versions of \qwen{}, \qwenbig{} and \microsoftphi{}. For the strong baselines, we select models across literature that have been aligned for polymer or chemistry domain: \Llamole{} ~\cite{llamaole}, \texttt{Llama-3.1-8B} based model, combining text and graph generation for molecular design and retrosynthesis using graph diffusion transformers; \LlamaSmol{} ~\cite{llasmol}, \texttt{Llama-2-7B} based LLM trained on \texttt{SMolInstruct} dataset for chemistry tasks for small molecules (molecular understanding, property predictions \& synthesis); \etherzero{} ~\cite{ether0}, a 24B chemistry-specialized LLM (\texttt{Mistral-Small-24B based)} fine-tuned with reinforcement learning, that uses Chain-of-thought for QA across a range of tasks (designing, completing, modifying, or synthesizing molecules); \ChemLLM{} ~\cite{chemllm} a \texttt{InternLM-2} based, QA model, instruction-tuned with ChemData for tasks such as Name Conversion, Property Prediction and so on.

\subsection{\dataset{} Evaluation Task Details}
\label{app:evaluation_task_details}

We evaluate \dataset{}-trained models on the same six task categories as the training set (detailed descriptions in Appendix~\ref{app:task_design_details}). The val and test sets are intentionally out-of-distribution with respect to the train set, containing unseen polymers, properties, and novel instructions to simulate real-world polymer discovery.

The evaluation spans multiple NLP problem types that test different model capabilities: (i) \textbf{open-ended QA} for conceptual and mechanistic reasoning, (ii) \textbf{numerical regression} for property prediction, (iii) \textbf{multiple-choice selection} for comparison and ranking under constraints, (iv) \textbf{structured generation} of \smiles{} strings for synthesis and design, and (v) \textbf{ranking} for pairwise property comparison. This coverage ensures \dataset{} tests not just domain knowledge, but also a model's ability to handle symbolic (\smiles{}), numerical, and free-form natural-language outputs -- formats that polymer researchers actually use in practice.

\subsection{External Benchmarks}
\label{app:external_benchmarks}

To assess generalization beyond types of queries present in \dataset{} and validate our models' ability to generalize to various polymer tasks, we evaluate on external benchmarks not seen during training. These datasets test different aspects of polymer understanding, and provide independent validations of our approach.

\paragraph{Block Polymers} ~\cite{jablonkagroup_block_polymers_morphology} This dataset contains questions about block copolymers, and their phase behavior. We use 2 subsets of this data for evaluation - \textit{Subset i0} that requires phase state predictions given polymer characteristics; and \textit{Subset i2} that requires inverse design i.e. generating polymer structures that achieve desired phase behaviour under specified constraints - both tasks involving OOD concepts w.r.t. \dataset{}.

\paragraph{Llamole} ~\cite{llamaole} evaluates molecular design and retro-synthesis capabilities requiring models to design molecules that satisfy multiple structural constraints (such as molecular weight, ring count, and rotatable bonds), and propose detailed synthesis routes with reagents and reaction conditions. This benchmark is particularly challenging as it requires generating complete synthesis procedures - a task beyond the scope of \dataset{}'s training. 

\paragraph{ChemData} ~\cite{chemllm} is a broad chemistry benchmark covering molecular understanding, classification, property prediction, and reasoning across diverse chemical domains. We select polymer-focused questions from this benchmark. This subset spans across a wide range of questions from simple classification (eg. identify whether a molecule is a polymer) to conceptual understanding (such as explaining polymerization mechanisms) and application-oriented reasoning (such as improving polymer membrane performance for water treatment). This dataset tests whether models can handle diverse polymer questions, presented in different formats and contexts - evaluating both breadth of polymer knowledge and robustness to question phrasing variations beyond \dataset{}.

\paragraph{PolyReal} ~\cite{liu2026polyrealbenchmarkrealworldpolymer} is a recently introduced benchmark for real world scientific tasks to evaluate Multimodal LLMs on the full lifecucle of a polymer experimentation including capabilities like data extraction, lab safety, and other practical aspects in Polymer Science. We use the text-only subset of the dataset as an external benchmark to assess our \dataset{}-trained models. While PolyReal is also focused on polymer science, it emphasizes practical scientific workflows and broad experimental reasoning, it serves as a complementary test of model generalization beyond the design-related tasks \dataset{} emphasizes on.

We remove any overlapping polymers between \dataset{} and the external datasets. We then randomly sample 200 questions each from \textit{Llamole} and \textit{ChemData}, and 200 questions from each of the two \textit{Block Polymers} subsets (400 total). For PolyReal, we use the text-only subset for evaluation (180 questions). In total, we use 980 questions for external evaluation.

\subsection{Evaluation Metrics (Detailed)}
\label{app:eval_metrics_detailed}

We employ task-specific quantitative and qualitative metrics to comprehensively assess model performance across diverse reasoning and generation tasks. Below we describe each metric in detail.

\paragraph{ROUGE-L (RgL).} Measures longest common subsequence between predicted and reference text~\cite{lin-2004-rouge}; higher indicates closer alignment.

\paragraph{LLM-as-a-Judge (Jdg)}  
Following \citet{zheng2023judging}, we employ a large language model (\gptOSS{}) to score model outputs on a \textit{Likert scale (1–7)} under a blind evaluation setup. This metric complements automatic scores by reflecting human-aligned preferences for coherence, factuality, and reasoning quality. Scores are averaged across annotators and instances to yield the final preference rating.

\paragraph{Pearson Correlation Coefficient (r)}  
The Pearson correlation coefficient quantifies linear agreement between predicted and ground-truth numeric values, defined as:
\[
r = \frac{\sum_i (x_i - \bar{x})(y_i - \bar{y})}{\sqrt{\sum_i (x_i - \bar{x})^2 \sum_i (y_i - \bar{y})^2}},
\]
where \(x_i\) and \(y_i\) are predicted and true values. A value close to 1 indicates strong positive correlation and thus better predictive consistency.

\paragraph{Mean Absolute Error (MAE)}  
MAE measures the average absolute deviation between predicted and reference values:
\[
\text{MAE} = \frac{1}{N}\sum_i |x_i - y_i|.
\]
Lower MAE indicates more accurate numerical predictions, complementing correlation-based metrics by reflecting absolute deviation magnitude.

\paragraph{Exact Match (EM)}  
For counting and discrete-output tasks, EM measures the percentage of predictions that exactly match the gold label. It is a strict metric emphasizing precise numerical or categorical correctness.

\paragraph{Similarity (Sim.)}  
For \smiles{} generation tasks, \texttt{Similarity} is computed as the Tanimoto coefficient between molecular fingerprints of the predicted and reference structures, using the \texttt{RDKit} library ~\cite{rdkit}. This metric evaluates whether generated molecules preserve the intended chemical and structural characteristics of the target compound, reflecting the model’s ability to reproduce meaningful molecular features.

\paragraph{Validity (Val.)}  
\texttt{Validity} measures the proportion of syntactically and chemically valid \smiles{} strings among generated samples, determined via \texttt{RDKit} parsing and sanitization routines ~\cite{rdkit}. It ensures that model outputs correspond to chemically plausible molecules adhering to \smiles{} grammar and chemical valency rules. High validity indicates robust chemical reasoning and structural correctness.

\paragraph{Synthetic Accessibility (SA) Scores}
\texttt{SA} scores measure the pratical difficulty of synthesizing generated molecules using the method of~\cite{sa-scores}. We compute SA scores for each predicted molecule using \texttt{RDKit}~\cite{rdkit}'s implementation of~\cite{sa-scores} algorithm \footnote{\url{https://github.com/rdkit/rdkit/blob/master/Contrib/SA_Score/sascorer.py}}; for multi-component outputs, we report the mean score across components. Lower scores indicate easier synthesis.

\paragraph{Accuracy (MCQ)}  
For multiple-choice question (MCQ) tasks, accuracy reflects the proportion of correctly selected options relative to the total number of questions, directly measuring discriminative reasoning performance.

\paragraph{Kendall’s Tau Correlation (Pair)}  
For pairwise ranking and comparison tasks, Kendall’s Tau measures ordinal agreement between predicted and reference rankings. It is computed as:
\[
\tau = \frac{C - D}{C + D},
\]
where \(C\) and \(D\) denote the number of concordant and discordant pairs, respectively. Values closer to 1 indicate strong agreement in ranking order.

Together, these metrics provide a balanced evaluation of factual accuracy, structural fidelity, and human-aligned reasoning quality across both symbolic and natural-language outputs. Additionally, we note that SMILES similarity and validity are coarse automated measures — they indicate preservation of structural motifs and chemical grammar compliance but do not guarantee laboratory synthesizability.


\subsection{Model Configuration}

All models are trained using LoRA (Low-Rank Adaptation)~\cite{hu2022lora} with the following configuration:
\begin{itemize}
    \item LoRA rank ($r$): 16
    \item LoRA alpha ($\alpha$): 32
    \item LoRA dropout: 0.05
    \item Target modules: all attention projections (Q, K, V, O) and MLP layers (gate, up, down projections)
    \item Precision: bfloat16
\end{itemize}


\subsection{Supervised Fine-Tuning (SFT) Hyperparameters}
\label{app:hyperparams}

Table \ref{tab:sft_hyperparams} presents the complete set of hyperparameters used during the instruction tuning phase.

\begin{table}[h]
\small
\centering
\begin{tabular}{ll}
\toprule
\textbf{Hyperparameter} & \textbf{Value} \\
\midrule
Learning rate & 1e-4* \\
Optimizer & AdamW \\
Per-device batch size & 8 \\
Gradient accumulation steps & 4 \\
Effective batch size & 32* \\
Number of epochs & 3 \\
Maximum sequence length & 512* \\
Learning rate schedule & Linear with warmup \\
Gradient checkpointing & Enabled \\
Mixed precision & bfloat16 \\
\bottomrule
\end{tabular}
\caption{Supervised fine-tuning hyperparameters used during instruction tuning. Parameters marked with (*) vary across model sizes and task settings. Learning rates were selected from \{1e-4, 5e-5\}, effective batch sizes ranged from 32--64, and maximum sequence lengths ranged from 512--1024 tokens.}
\label{tab:sft_hyperparams}
\end{table}

\paragraph{Custom Tokens.}
During supervised fine-tuning of \dataset{} models, we introduce a set of task-specific control tokens to explicitly mark structured chemical and reasoning spans: 
\texttt{<iupac\_start>, <iupac\_end>} for IUPAC names, 
\texttt{<smiles\_start>, <smiles\_end>} for SMILES strings, 
\texttt{<answer\_start>, <answer\_end>} for answer boundaries, and 
\texttt{<think\_start>, <think\_end>} for chain-of-thought reasoning traces. 
These tokens provide consistent delimiters that help the model learn to separate molecular representations, final answers, and intermediate reasoning steps during supervised fine-tuning.

\subsection{Computational Resources}
All experiments are conducted using NVIDIA A100 GPUs with the Hugging Face Transformers library and TRL (Transformer Reinforcement Learning) framework. Flash Attention 2 is optionally enabled for improved efficiency.

\subsection{Generation Configuration}

Generation is performed in both zero-shot and few-shot settings using nucleus sampling across temperatures ${0.1, 0.5, 0.7}$ with up to five diverse outputs (\texttt{num\_return\_sequences=5}). We report the best-performing scores per model across these settings.

\section{Results \& Analysis}
\label{app:results_detailed}

\subsection{Additional Findings}
\label{app:smiles_findings}

\begin{table}[t]
\centering
\small
\begin{tabular}{lccc}
\toprule
Model & Sim. $\uparrow$ & Val. $\uparrow$ & SA $\downarrow$ \\
\midrule
GPT-4o & 0.16 & 0.11 & 3.92 \\
Claude 4.6 & 0.37 & \textbf{0.93} & 5.16 \\
GPT-5.4 & 0.28 & 0.58 & 4.87 \\
\midrule
Qwen32B + \dataset{} & 0.64 & 0.92 & 3.47 \\
Qwen14B + \dataset{} & 0.72 & 0.94 & 3.41 \\
ChemLLM + \dataset{} & 0.74 & \textbf{0.95} & \textbf{3.04} \\
Phi14B + \dataset{} & \textbf{0.82} & 0.66 & 3.42 \\
\bottomrule
\end{tabular}
\caption{Synthetic accessibility analysis for SMILES generation (Task 6). Higher is better for Similarity and Validity; lower is better for Synthetic Accessibility (SA).}
\label{tab:sa_scores}
\end{table}

\paragraph{Strong SMILES Representation Performance.}
\dataset{}’s comprehensive SMILES coverage is a key factor behind its gains in Design and Synthesis tasks. 
As shown in Table~\ref{tab:main_results}, \dataset{}-tuned models achieve substantial improvements in both structural \textit{similarity} and \textit{validity} metrics—nearly tripling similarity scores over baselines. This suggests that \dataset{} effectively teaches models chemically consistent SMILES syntax and polymer-specific bonding patterns.
To access whether the practical synthesizability of the generated \smiles{} as part of the design tasks, we additionally evaluate \textit{Synthetic Accessibility (SA)} scores~\cite{sa-scores}. Lower SA scires indicate easier synthesis, values in ranges 1--3 are considered readily synthesizable, while scores above 6 are considered difficult to synthesize. Table~\ref{tab:sa_scores} shows while \dataset{}-trained models outperform frontier models in \texttt{Tanimoto-similarity}, they maintain comparable or lower SA scores.  

Together, these results suggest that SMILES learning is not just a data-format advantage but a key enabler of polymer design reasoning.

\subsection{Compositionality Gap Analysis Details}
\label{app:additional_diagnostics}
We probe whether errors in polymer design arise from missing atomic skills or from failures to select and compose otherwise-correct subskills into a constraint-satisfying solution \cite{press-etal-2023-measuring}. We use a sample of 300 questions from \dataset{}'s validation set, and prompt \gptOSS{} to decompose each question into 3--15 hierarchical sub-questions (sub-Qs) along with a set of required skills to answer each sub-question, which serve as our diagnostic probes to isolate the root cause of the issue that leads to low scores. In total, we have a total of 2.6K subquestions. We generate "gold" answers for these sub-questions using \gptFourO{} to be used in our experiments as context.  

Our investigation is structured around three core experiments:

\begin{itemize}[leftmargin=*] \item \textbf{Exp 1: The Skill Gap (Atomic Knowledge):} We evaluate whether the model can answer the sub-Qs in isolation. This tests the baseline "atomic" knowledge (e.g., monomer molecular weights, functional group identification) required for the task. 

\item \textbf{Exp 2: Skill Selection \& Evaluation:} We evaluate the model's self-generated Chain-of-Thought (CoT). We measure \textbf{Recall} (did the model's CoT mention the necessary sub-Qs?) and \textbf{Precision} (were the mentioned sub-steps actually correct?). 

\item \textbf{Exp 3: The Compositional Gap:} We provide the model with "Gold" answers to the sub-Qs as thinking guidance before it generates the final polymer design. If performance doesn't improve significantly, it confirms the failure was not a lack of knowledge, but a failure in integrating all the information into a coherent, structured narrative. \end{itemize}

\begin{table}[t]
\centering
\setlength{\tabcolsep}{6pt}
\begin{small}
\begin{tabular}{l|c|c}
\hline
\textbf{Skill} & \textbf{Precision (\%)} & \textbf{\% of Questions} \\
\hline
\texttt{DEFINITION}   & 89.25 & 22.00 \\
\texttt{STRUCTURE}    & 72.71 & 27.58 \\
\texttt{RULES}        & 96.67 & 0.19  \\
\texttt{COUNTING}     & 72.58 & 1.19  \\
\texttt{NOMENCLATURE} & 83.33 & 1.27  \\
\texttt{COMPARISON}   & 83.74 & 6.23  \\
\texttt{EXPERIMENTAL} & 81.94 & 12.92 \\
\texttt{INFERENCE}    & 89.43 & 29.12 \\
\hline
\end{tabular}
\end{small}
\caption{Experiment 1: The Skill Gap - Mean Precision by core skill category. \% of Questions computed out of 2.6K total questions.}
\label{tab:exp1_precision_by_skill}
\end{table}

\begin{table}[t]
\centering
\setlength{\tabcolsep}{5pt}
\footnotesize
\begin{tabular}{l|c|c}
\hline
\textbf{Task Cat.} & \textbf{Precision (\%)} & \textbf{Recall (\%)} \\
\hline
Structural Understanding & 79.5 & 47.2 \\
Conceptual Knowledge & 93.0 & 47.4 \\
Property Prediction & 44.0 & 16.4 \\
Comparison \& Ranking & 11.3 & 7.0 \\
Adv. Property Reasoning & 84.8 & 44.1 \\
Synthesis \& Design & 36.3 & 16.6 \\
\hline
\end{tabular}
\caption{Experiment 2: Precision and recall by PolyBench task category.}
\label{tab:exp2_precision_recall_by_task_category}
\end{table}

\begin{table}[t]
\centering
\footnotesize
\setlength{\tabcolsep}{4pt}
\begin{tabular}{lccc}
\toprule
\textbf{Task Category} & \textbf{Main} & \textbf{SubQs} & \textbf{SubQs+A} \\
\midrule
Structural Understanding    & 5.08 & 4.68 & 6.02 \\
Conceptual Knowledge        & 4.98 & 6.26 & 6.48 \\
Property Prediction         & 3.14 & 2.96 & 3.46 \\
Comparison \& Ranking       & 4.72 & 4.49 & 5.92 \\
Adv.\ Property Reasoning    & 5.10 & 5.94 & 6.46 \\
Synthesis \& Design         & 4.30 & 4.84 & 5.28 \\
\bottomrule
\end{tabular}
\caption{Mean scores by PolyBench task category for the main setting (no extra context), SubQs-only context, and SubQs + answers context. Higher is better.}
\label{tab:exp3_merged_scores_by_task}
\end{table}

\paragraph{Key Findings.}

Exp 1 (Table \ref{tab:exp1_precision_by_skill}) shows that the model answers many sub-questions correctly in isolation, with high mean precision across several skill types (e.g., \textsc{Definition} and \textsc{Inference} $\approx$89\%). Errors are more concentrated in structure-heavy skills (\textsc{Structure}/\textsc{Counting} $\approx$72\%), suggesting that failures are not purely due to missing factual knowledge but also arise from harder structural manipulations. This is in line with our evaluation on \dataset{} in Table \ref{tab:main_results} where we see the model struggle with \taskOnestrunderstanding{}.

Exp~2 (Table \ref{tab:exp2_precision_recall_by_task_category}) reveals that self-generated reasoning traces frequently miss required sub-goals: while precision is often high for some categories (e.g., Conceptual Knowledge 93\%, Advanced Property Reasoning 84.8\%), recall remains low (typically $\sim$44--47\% even for strong categories, and 7\% for Comparison/Ranking), suggesting that important constraints and intermediate steps are omitted - especially for constraint-heavy decision tasks and property prediction. 

Finally, Exp~3 (Table \ref{tab:exp3_merged_scores_by_task}) shows that providing sub-questions alone yields mixed effects, whereas providing \emph{gold} sub-question answers generally improves scores (largest gains in Structural Understanding and Comparison/Ranking), confirming that missing atomic knowledge is not the only bottleneck; rather, even with correct components available, integrating multiple constraints and signals into a coherent final decision remains challenging, consistent with a compositionality gap.

\subsection{Analysis Details: \texttt{LLM-as-a-Judge}}
\label{app:llm_as_judge_analysis}

This section provides additional details for the \texttt{LLM-as-a-Judge} evaluation framework, including judge-human agreement statistics across \dataset{} and external datasets, category-wise length-bias correlations with $p$-values, and per-score length breakdowns.

\begin{table}[h]
\centering
\footnotesize
\begin{tabular}{lc}
\toprule
\textbf{Model} & \textbf{Corr.} \\
\midrule
\microsoftphi{} + CoT    & 0.11* \\
\qwenbig{} + CoT & 0.03\phantom{*} \\
\gptfive{}        & 0.15* \\
\bottomrule
\end{tabular}
\caption{Pearson correlation between prediction length and LLM-as-a-Judge score (all tasks). * denotes p < 0.05. Correlations are weak; Qwen-2.5 shows no significant length–score relationship.}
\label{tab:length-score-overall}
\vspace{-1em}
\end{table}

We use \texttt{LLM-as-a-Judge} to evaluate open-ended QA questions in \dataset{} and external datasets; for all other question types, we rely on deterministic metrics (exact match, SMILES similarity, Kendall's $\tau$ pairwise accuracy, etc.). For open-ended QA, we additionally report ROUGE-L, and our findings show that trends across the two metrics consistently agree. To further validate our use of \texttt{LLM-as-a-Judge}, we conduct two analyses: (a) calibrating judge scores against human expert ratings, and (b) empirically probing for length and stylistic bias by comparing models with distinct output styles.

\subsubsection{Judge Reliability Assessment}
\label{app:judge_reliability}

We compare human judgments against \texttt{LLM-as-a-Judge} scores on 250 questions sampled from our test sets. Human annotators rate each response on a 5-point Likert scale for factual accuracy, reasoning soundness, and overall alignment with the reference solution. Table~\ref{tab:llm_judge_validation_combined} reports Spearman $\rho$ and Kendall $\tau$ for rank-order agreement, along with \textit{Bias}: the mean score difference between human and LLM ratings (Human $-$ LLM). A positive bias means humans rate responses higher than the judge on average -- indicating the judge is strict relative to humans; a negative bias indicates the judge is lenient. Results are broken down by dataset (\dataset{} Test and External) and by the source model whose outputs were scored (\microsoftphi{}+CoT, \gptFourO{}, \claudeSonnet{}). Agreement is consistently higher on \dataset{} (Test) than on External, and bias is small and positive on \dataset{} (Test) but skews negative on External -- particularly for \gptFourO{} outputs, where humans rate responses more than a full point lower than the judge does.

\begin{table}[h]
\centering
\scriptsize
\begin{tabular}{llccr}
\toprule
\textbf{Dataset} & \textbf{Model} & \textbf{$\rho$} & \textbf{$\tau$} & \textbf{Bias} \\
\midrule
\multirow{3}{*}{\dataset{} (Test)}
  & \sonnet{}          & 0.792 & 0.723 & $+$0.23 \\
  & Phi-4+CoT (Ours)   & 0.760 & 0.678 & $+$0.21 \\
  & \gptFourO{}        & 0.574 & 0.486 & $-$0.21 \\
\midrule
\multirow{3}{*}{External}
  & Phi-4+CoT (Ours)   & 0.660 & 0.561 & $-$0.43 \\
  & \gptFourO{}        & 0.487 & 0.411 & $-$1.14 \\
  & \sonnet{}          & 0.488 & 0.472 & $-$0.01 \\
\bottomrule
\end{tabular}
\caption{\texttt{LLM-as-a-Judge} validation against human ratings. \gptOSS{} was re-run at 1--5 on the calibration subset to match the 5-point human scale. Spearman $\rho$ and Kendall $\tau$ indicate rank-order agreement; Bias: mean score difference (Human $-$ LLM) on a 5-point scale; positive bias indicates humans rate higher than the judge. All correlations are statistically significant ($p < 0.0001$).}
\label{tab:llm_judge_validation_combined}
\end{table}

\subsubsection{Length and Stylistic Bias}
\label{app:length_bias}
\begin{figure*}[t]
\centering
\includegraphics[width=0.95\textwidth]{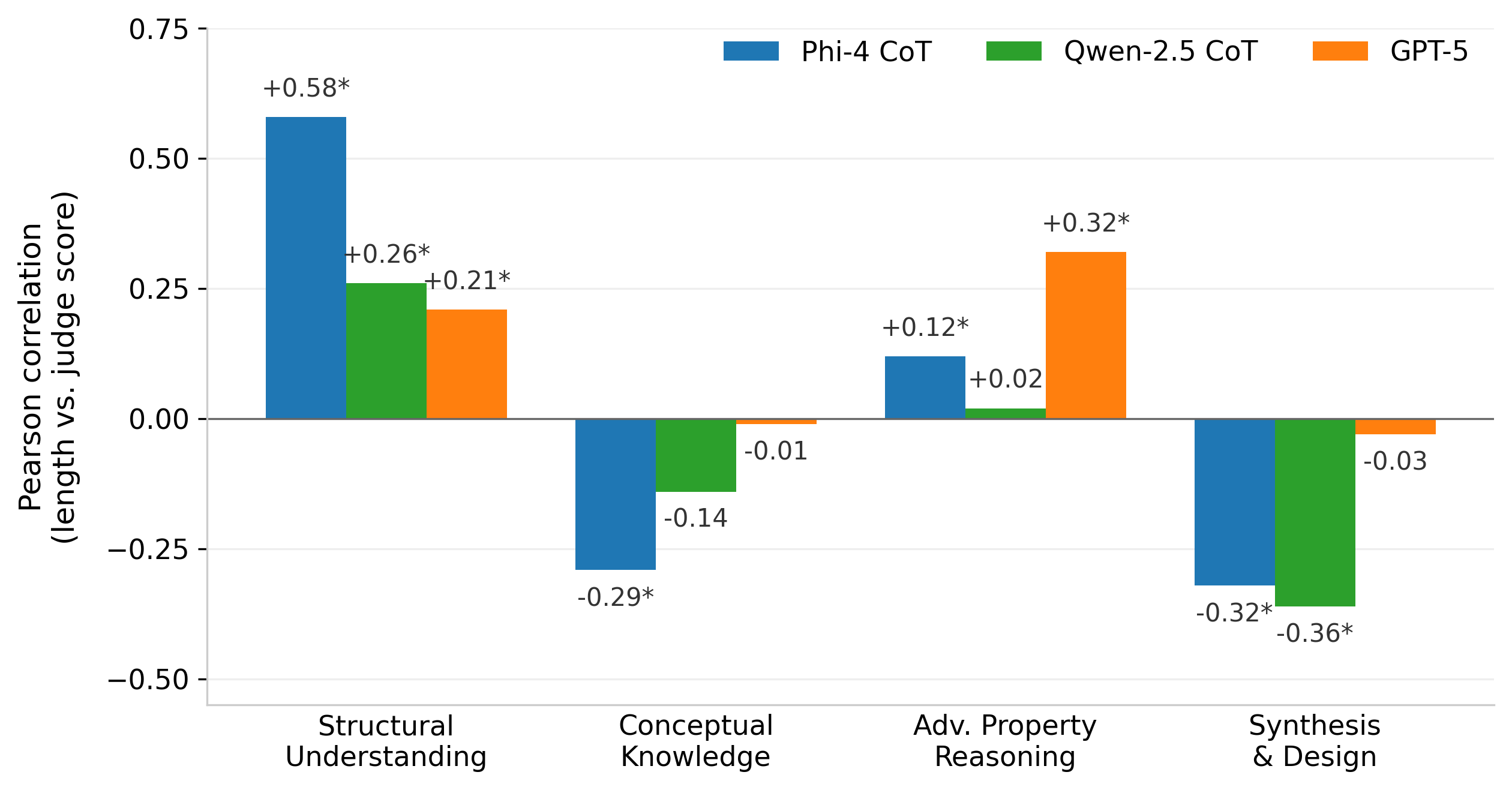}
\caption{\textbf{Category-wise length--score correlations by model.} Pearson correlation between prediction length and \texttt{LLM-as-a-Judge} score across four task categories for three models: \microsoftphi{}+CoT and \qwenbig{}+CoT are our \dataset{}-trained models, and \gptfive{} is a frontier closed-source baseline. * denotes $p < 0.05$.}
\label{fig:length_bias}
\end{figure*}

We analyze length-driven bias across different model outputs -- \dataset{}-trained models (\microsoftphi{}+CoT and \qwenbig{}+CoT) and \gptfive{}. Overall length--score correlations are weak across all models (Pearson $r \leq 0.15$; Table~\ref{tab:length-score-overall}). Task-wise correlations (Figure~\ref{fig:length_bias}) flip sign within the same category across models, inconsistent with a uniform preference for longer outputs or for any fixed output style. The score--length relationship is also non-monotonic (Table~\ref{tab:judge-length-by-score}): top-rated predictions (score 7) are shorter than score-6 outputs, and \gptfive{}'s verbose responses (averaging 2--4$\times$ the length of \dataset{}-trained outputs) do not translate into higher scores. Together, these results indicate that verbosity and stylistic bias have a limited effect on our evaluation (more details in Appendix~\ref{app:llm_as_judge_analysis}).


\begin{table}[h]
\centering
\small
\begin{tabular}{cccc}
\toprule
\textbf{Judge Score} & \textbf{Phi-4 CoT} & \textbf{Qwen-2.5 CoT} & \textbf{GPT-5} \\
\midrule
1 & \phantom{00}4.51 & \phantom{00}6.73 & \phantom{00}6.61 \\
2 & \phantom{0}45.10 & \phantom{0}35.43 & \phantom{0}33.88 \\
3 & \phantom{0}65.76 & \phantom{0}50.00 & 102.57 \\
4 & \phantom{0}63.02 & \phantom{0}51.11 & \phantom{0}62.53 \\
5 & \phantom{0}66.45 & \phantom{0}64.48 & 254.52 \\
6 & \phantom{0}71.92 & \phantom{0}68.63 & 143.18 \\
7 & \phantom{0}37.22 & \phantom{0}28.25 & \phantom{0}91.53 \\
\bottomrule
\end{tabular}
\caption{Average prediction length (word count) at each \texttt{LLM-as-a-Judge} score level, by model.}
\label{tab:judge-length-by-score}
\end{table}

\subsection{Analysis Details: Quality of CoT}
\label{app:quality_of_COT_analysis}

Unlike math or programming, polymer reasoning has no oracle for scalable verification, making distillation one of the few practical mechanisms for scaling reasoning supervision in this domain. Our pipeline mitigates error propagation through three safeguards: (1) knowledge grounding, (2) automated fact-checking, and (3) manual validation by polymer scientists. In this section, we detail our manual validation and automated fact-checking, along with a characterization of residual errors in the distilled traces.

\begin{figure}
    \centering
    \includegraphics[width=1\linewidth]{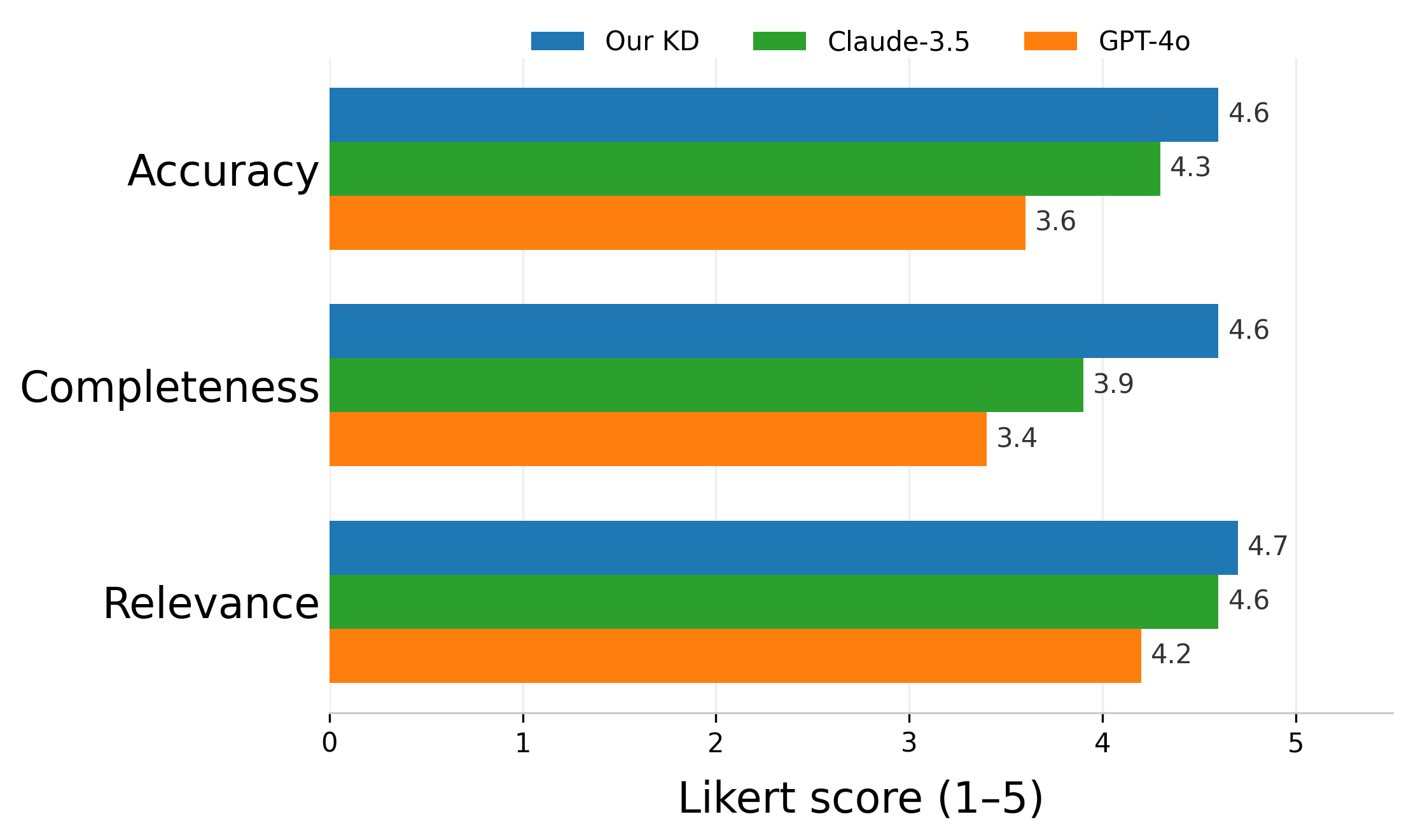}
    \caption{CoT quality ratings of our KD Pipeline against \gptFourO{} and \sonnet{} for Accuracy, Completeness, and Relevance, on a 1--5 Likert scale by human experts.}
\label{fig:cot_quality}
\vspace{-1.5em}
\end{figure}

\paragraph{Human Evaluation of CoT Traces}
We sample 100 questions from the train set of \dataset{} and prompt \gptFourO{} and \claudeSonnet{} to produce CoT responses -- without any instructions or additional information about the polymer -- comparing these against CoTs from our knowledge distillation pipeline. Annotators evaluate all three outputs (anonymized for source) on a 1--5 Likert scale across relevance, completeness, and accuracy.

\paragraph{Human Evaluation of Model Outputs}
We sample 250 questions from our test sets (\dataset{} and external) and collect outputs from \microsoftphi{}+CoT (our best \dataset{}-trained model), \gptFourO{}, and \claudeSonnet{}. Evaluators select their preferred response(s) from the three options — annotators may mark multiple responses when they are comparably strong, so per-model preference rates are reported independently and do not sum to 100\%; results are in Figure 5.

\begin{figure}
\centering
\includegraphics[width=1\linewidth]{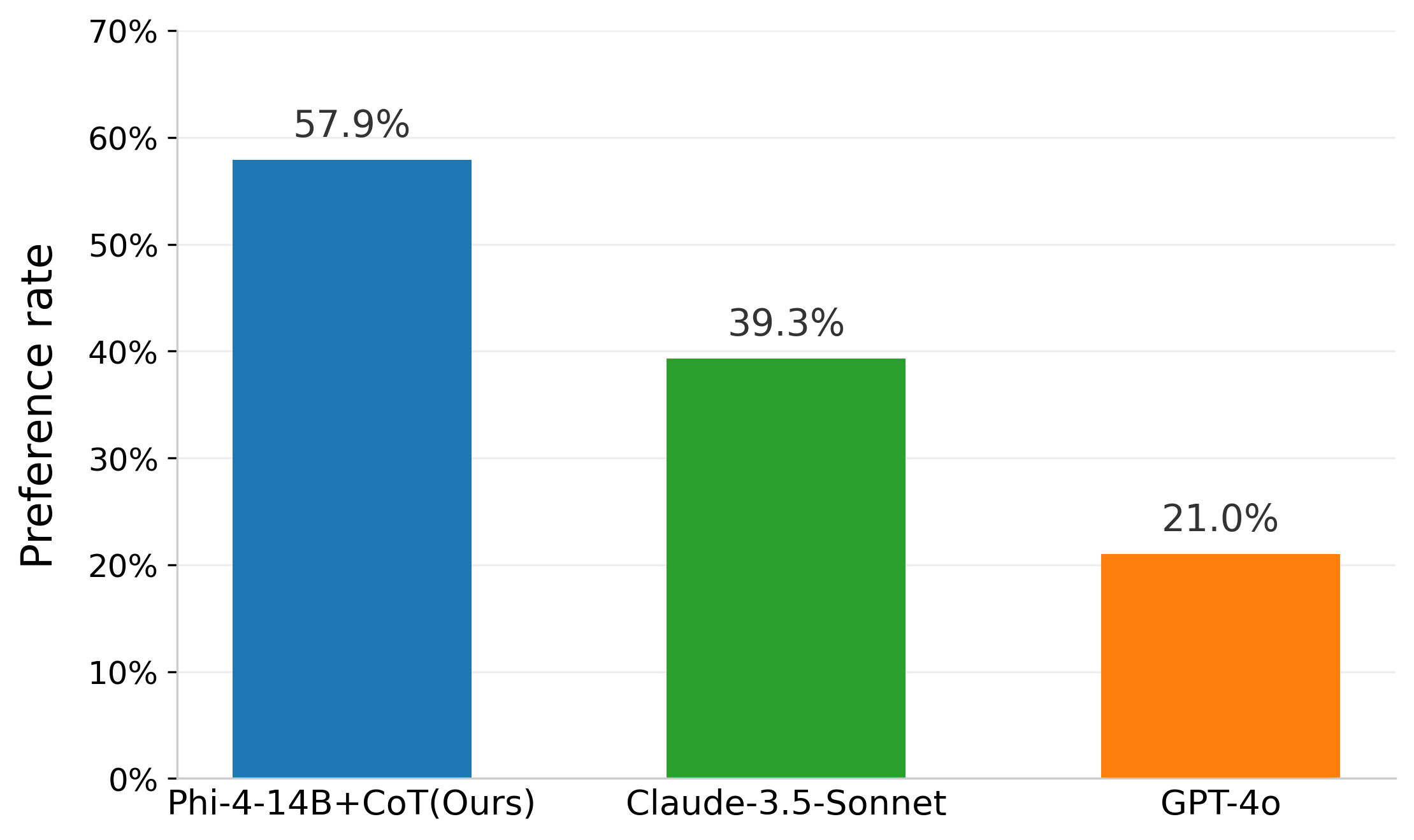}
\caption{\textbf{Human preference study on model outputs.} Blind annotators select the preferred response among \microsoftphi{}\texttt{+CoT}, \gptFourO{}, and \claudeSonnet{} on a held-out set of PolyBench questions. Annotators may select multiple responses per question when more than one is comparably strong, so rates do not sum to 100\%. Preference rates indicate that \microsoftphi{}\texttt{+CoT} is chosen more often overall, supporting the automated evaluation trends.}
\label{fig:human_eval_pref}
\end{figure}

Annotators rate our knowledge-augmented CoT traces as more complete, accurate, and relevant than teacher-only CoTs (Figure~\ref{fig:cot_quality}), and prefer \dataset{}-trained model outputs over \gptFourO{} and \claudeSonnet{} (Figure~\ref{fig:human_eval_pref}). Both results corroborate our automated evaluations.

\paragraph{Automatic Verification Pipeline \& Error Characterization} On a stratified sample of 300 (train+val) CoT traces, 80.67\% are flagged as fully correct (score 5 on a 1--5 Likert scale). Further analysis of the remaining 19.33\% shows the dominant errors (Table~\ref{tab:cot-errors}) are missing steps and incomplete reasoning -- e.g., oversimplification or skipped intermediate steps. These errors are confined to the CoT; final answers are ground-truth labels.
We evaluate CoT quality using \gptfive{} as a judge, prompting it to score each trace on a 1--5 Likert scale and identify errors from a fixed taxonomy of 12 categories: logical error, arithmetic error, hallucination, contradiction, irrelevant reasoning, missing steps, invalid assumption, misinterpretation of question, answer mismatch, incomplete reasoning, strategy misuse, and unclear explanation. For each QA instance, the judge receives the question, gold answer, and distilled CoT, and returns a JSON object with a quality score and list of error labels (Table~\ref{tab:cot-quality}).

Representative flagged traces are shown in Table~\ref{tab:cot-error-examples}.

These analyses suggest limited error propagation from distillation, and is supported by our downstream results: \dataset{}-trained models outperform their base variants and sometimes closed-source models (Table~\ref{tab:main_results}), transfer well to external benchmarks (Table~\ref{tab:external_results_with_pass5}), and are preferred 57.9\% of the time over frontier models (Figure~\ref{fig:human_eval_pref}) -- performance patterns unlikely under heavy teacher contamination.

\begin{table}[h]
\centering
\small
\begin{tabular}{llcc}
\toprule
\textbf{CoT Quality} & \textbf{Criteria} & \textbf{Count} & \textbf{\%} \\
\midrule
Good & Score=5 & 242 & 80.67 \\
Bad  & Score<5 & \phantom{0}58 & 19.33 \\
\bottomrule
\end{tabular}
\caption{Automatic verification of CoT quality on 300 stratified training-set traces.}
\label{tab:cot-quality}
\end{table}

\begin{table}[h]
\centering
\footnotesize
\begin{tabular}{lcc}
\toprule
\textbf{Error Type} & \textbf{Count} & \textbf{\%} \\
\midrule
Missing steps         & 19 & 32.7 \\
Incomplete reasoning  & 16 & 27.5 \\
Unclear explanation   & \phantom{0}6 & 10.3 \\
Irrelevant reasoning  & \phantom{0}4 & 6.9 \\
\bottomrule
\end{tabular}
\caption{Top 4 error types among the 19.33\% flagged as incorrect. Percentages are of the flagged subset.}
\label{tab:cot-errors}
\vspace{-1em}
\end{table}

\begin{table*}[h]
\centering
\footnotesize
\begin{tabular}{p{0.03\linewidth} p{0.20\linewidth} p{0.20\linewidth} p{0.22\linewidth} p{0.20\linewidth}}
\toprule
\textbf{Score} & \textbf{Error labels} & \textbf{Question} & \textbf{Chain-of-Thought} & \textbf{Issue} \\
\midrule

4 & \texttt{missing\_steps}, \texttt{unclear\_explanation}
& In one repeat unit of poly(oxynaphthalene-1,5-diyloxycarbonyl-1,2-phenyleneethene[...]), how many oxygen atoms are present? \textit{Gold: 4.}
& There are two ether linkages [...] two carbonyl groups [...] Total = 4.
& Skips explicit identification of functional groups in the full structure and does not map each oxygen to specific subunits, making the derivation hard to verify. \\
\midrule

2 & \texttt{missing\_steps}, \texttt{incomplete\_reasoning}
& Design a polymer with thermal stability $>$300\textdegree C using biphenyl units [...] cyano groups [...] meta-substitution. \textit{Gold:} polymer satisfying all four stated constraints.
& Lists design requirements [...] A diamine--diacyl chloride condensation would form these linkages effectively.
& Enumerates what each feature \textit{should} do but never verifies the proposed monomers satisfy the 1,2-biphenyl or meta-substitution constraints. \\
\midrule

3 & \texttt{unclear\_explanation}, \texttt{incomplete\_reasoning}
& How do ortho amino groups and aromatic core in benzene-1,2-diamine contribute to its bandgap of 3.7\,eV? \textit{Gold:} amine lone pairs donate into $\pi$ system [...] setting HOMO--LUMO gap.
& Ortho amino groups ``create localized electron density'' [...] ``generate specific orbital overlap geometries'' [...] combination creates ``specific electronic effects.''
& Vague phrases without naming the actual electronic mechanism or connecting to 3.7\,eV. \\
\midrule

3 & \texttt{irrelevant\_reasoning}, \texttt{incomplete\_reasoning}
& \texttt{Cc1cc(C2CO2)ccc1Cl} shows high amine reactivity but poor solvent resistance [...] what structural features are responsible? \textit{Gold:} reactive epoxy $+$ Cl for reactivity [...] insufficient rigidity for solvent resistance.
& Epoxy undergoes ring-opening [...] Cl makes epoxy more electrophilic [...] poor solvent resistance suggests insufficient crosslinking [...] methyl reducing chain packing.
& Invokes ``insufficient crosslinking'' for a monomer-level question (crosslinking depends on polymerization, not monomer structure), drifts to irrelevant factors. \\

\bottomrule
\end{tabular}
\caption{Representative CoT traces flagged by the automatic verification pipeline, covering the major CoT error types. Score is the judge's 1--5 quality rating}

\label{tab:cot-error-examples}
\end{table*}

\section{Prompts}

\subsection{Prompt for Task Generation}
\label{app:qa_prompt}

For open-ended QA tasks in \dataset{} (\taskOnestrunderstanding{}, \taskTwopolymerconcepts{}, and \taskFivereasoning{}), we automatically generate compositional question--answer pairs using the prompt shown in Figure~\ref{fig:prompt_task_generation}. The prompt conditions question generation on each polymer's structural and experimental profile, producing self-contained questions with chemically grounded answers.

\vspace{2pt}

\begin{figure*}[t]
\centering
\begin{tcolorbox}[
    width=\textwidth,
    title=\textbf{Task Generation Prompt (Knowledge-Augmented)},
    colback=gray!5,
    colframe=black!50,
    fonttitle=\bfseries,
    before upper=\ttfamily\small
]
\#\#\# Instruction:\\
Generate exactly \{\} compositional questions for the specified category based on the provided polymer description paragraph.
Questions must be completely self-contained with all necessary inputs included within the question itself,
use specific polymer/monomer names with \texttt{<iupac\_start><iupac\_end>} and \texttt{<smiles\_start><smiles\_end>} tags,
and employ natural language.
Answers should provide detailed reasoning based solely on the structural and property information given in each question,
without referencing the polymer description paragraph or external sources.
Output format: JSON with numbered question-answer pairs for the specified category.\\
\#\#\# Category:\\
\{\}\\
\#\#\# Category Description:\\
\{\}\\
\#\#\# Example Question-Answer for this Category:\\
\{\\
\hspace*{1em}"1": \{\\
\hspace*{2em}"question": "\{\}",\\
\hspace*{2em}"answer": "\{\}"\\
\hspace*{1em}\}\\
\}\\
\#\#\# Polymer Description:\\
\{Polymer Identifiers\}\\
\{Structural and Experimental Data\}\\
\#\#\# Output:
\end{tcolorbox}
\caption{Prompt template used by \claudeSonnet{} for generating open-ended QA tasks in \dataset{}. The prompt grounds question-generation in a specific polymer's identifiers, structural features, and experimental properties, producing category-appropriate QA pairs with reasoning.}
\label{fig:prompt_task_generation}
\end{figure*}

\subsection{Prompts for Knowledge Distillation}
\label{app:cot_prompts}

We use the prompt in Figure~\ref{fig:prompt_cot_distillation} to generate Chain-of-Thought traces in our knowledge-augmented distillation pipeline. The prompt conditions reasoning generation on both the question and the complete polymer profile, encouraging mechanistic explanations grounded in the molecular and experimental data provided.

\begin{figure*}[t]
\centering
\begin{tcolorbox}[
    width=\textwidth,
    title=\textbf{Knowledge Distillation Prompt (CoT Generation)},
    colback=gray!5,
    colframe=black!50,
    fonttitle=\bfseries,
    before upper=\ttfamily\small
]
\#\#\# INSTRUCTION:\\
You are a chemistry reasoning expert. Analyze the given polymer and its experimental property value.
Provide a structured, step-by-step scientific explanation (Chain-of-Thought) describing why the polymer exhibits this value.
Focus on the relationships between molecular structure, intermolecular interactions, and resulting macroscopic properties.
Use clear, concise, and mechanistic reasoning grounded in chemistry.\\
---\\

\#\#\# QUESTION:\\
\{question\}\\

\#\#\# POLYMER DETAILS:\\
\{Polymer Details\}\\

\#\#\# OUTPUT:
\end{tcolorbox}
\caption{Prompt template used for Chain-of-Thought distillation of polymer property-based questions in our knowledge-augmented pipeline.}
\label{fig:prompt_cot_distillation}
\end{figure*}

\subsection{Prompts for Instruction Tuning}
\label{app:instruction_tuning_prompts}

We employ Alpaca-style templates~\cite{alpaca} for instruction-tuning of \dataset{} models, designed to handle both reasoning-augmented (with CoT) and direct-answer (without CoT) formulations. These templates enable the model to learn structured polymer reasoning while maintaining compatibility with common instruction-tuned formats. The CoT variant is shown in Figure~\ref{fig:prompt_sft_cot} and the No-CoT variant in Figure~\ref{fig:prompt_sft_nocot}.

\begin{figure*}[t]
\centering
\begin{tcolorbox}[
    width=\textwidth,
    title=\textbf{Instruction Tuning Prompt (with CoT)},
    colback=gray!5,
    colframe=black!50,
    fonttitle=\bfseries,
    before upper=\ttfamily\small
]
Below is an instruction that describes a task, paired with an input that provides further context. Write a response that appropriately completes the request.\\

\#\#\# Instruction:\\
\{\}\\

\#\#\# Thought:\\
\{\}\\

\#\#\# Response:\\
\{\}
\end{tcolorbox}
\caption{Instruction tuning prompt with Chain-of-Thought (CoT).}
\label{fig:prompt_sft_cot}
\end{figure*}

\begin{figure*}[t]
\centering
\begin{tcolorbox}[
    width=\textwidth,
    title=\textbf{Instruction Tuning Prompt (without CoT)},
    colback=gray!5,
    colframe=black!50,
    fonttitle=\bfseries,
    before upper=\ttfamily\small
]
Below is an instruction that describes a task, paired with an input that provides further context. Write a response that appropriately completes the request.\\

\#\#\# Instruction:\\
\{\}\\

\#\#\# Response:\\
\{\}
\end{tcolorbox}
\caption{Instruction tuning prompt without Chain-of-Thought (No-CoT).}
\label{fig:prompt_sft_nocot}
\end{figure*}

During supervised fine-tuning, both formats are applied based on data availability: samples with explicit reasoning use the CoT variant, while others use the No-CoT template. This separation ensures that \dataset{} models learn to perform both stepwise chemical reasoning and concise direct answering, depending on the task type.

\subsection{Prompts for Evaluation}
\label{app:instr_qa_dep}

We evaluate models under three prompting settings that vary along two dimensions: (i) whether in-context examples are provided, and (ii) whether the model is asked to produce explicit reasoning. The three settings are: (A) examples with reasoning (Figure~\ref{fig:prompt_examples_reasoning}), (B) examples without reasoning (Figure~\ref{fig:prompt_examples_no_reasoning}), and (C) no examples without reasoning (Figure~\ref{fig:prompt_no_examples_no_reasoning}).

\begin{figure*}[t]
\centering
\begin{tcolorbox}[
    width=\textwidth,
    title=\textbf{Setting A: Examples + Reasoning},
    colback=gray!5,
    colframe=black!50,
    fonttitle=\bfseries,
    before upper=\ttfamily\small
]
Below is an instruction that describes a task, paired with an input that provides further context.
Write a response that appropriately completes the request.\\

\#\#\# Instruction:\\
You are a helpful assistant for polymer chemistry. Think step-by-step before providing your final answer. Use example QA as a guide.\\

Examples:\\
\{\}\\

\#\#\# Question:\\
\{\}\\

\#\#\# Thought:\\
\{\}\\

\#\#\# Response:
\end{tcolorbox}
\caption{Evaluation prompt with in-context examples and explicit reasoning (CoT).}
\label{fig:prompt_examples_reasoning}
\end{figure*}

\begin{figure*}[t]
\centering
\begin{tcolorbox}[
    width=\textwidth,
    title=\textbf{Setting B: Examples + No Reasoning},
    colback=gray!5,
    colframe=black!50,
    fonttitle=\bfseries,
    before upper=\ttfamily\small
]
Below is an instruction that describes a task, paired with an input that provides further context.
Write a response that appropriately completes the request.\\

\#\#\# Instruction:\\
You are a helpful assistant for polymer chemistry. Provide only the final answer. Use example QA as a guide.\\

Examples:\\
\{\}\\

\#\#\# Question:\\
\{\}\\

\#\#\# Response:
\end{tcolorbox}
\caption{Evaluation prompt with in-context examples and no reasoning.}
\label{fig:prompt_examples_no_reasoning}
\end{figure*}

\begin{figure*}[t]
\centering
\begin{tcolorbox}[
    width=\textwidth,
    title=\textbf{Setting C: No Examples + No Reasoning},
    colback=gray!5,
    colframe=black!50,
    fonttitle=\bfseries,
    before upper=\ttfamily\small
]
Below is an instruction that describes a task, paired with an input that provides further context.
Write a response that appropriately completes the request.\\

\#\#\# Instruction:\\
You are a helpful assistant for polymer chemistry. Provide only the final answer.\\

\#\#\# Question:\\
\{\}\\

\#\#\# Response:
\end{tcolorbox}
\caption{Evaluation prompt with no in-context examples and no reasoning (zero-shot direct answer).}
\label{fig:prompt_no_examples_no_reasoning}
\end{figure*}

\subsection{Prompts for Automatic Evaluation (\texttt{LLM-as-a-Judge})}
\label{app:auto_eval_llm}

We use two versions of the \texttt{LLM-as-a-Judge} prompt depending on the evaluation task. For QA tasks in \dataset{} and external benchmarks, we use the 1--7 scale prompt (Figure~\ref{fig:prompt_llm_judge}); for CoT quality evaluation on distilled training traces, we use the 1--5 scale prompt (Figure~\ref{fig:prompt_llm_judge_5scale}). Both prompts include example outputs for all score levels (omitted here for space); a representative subset of scores (low, mid, high) is shown in each figure.

\begin{figure*}[t]
\centering
\begin{tcolorbox}[
    width=\textwidth,
    title=\textbf{\texttt{LLM-as-a-Judge} Prompt},
    colback=gray!5,
    colframe=black!50,
    fonttitle=\bfseries,
    before upper=\ttfamily\small
]
\#\#\# INSTRUCTION:\\
Evaluate the CANDIDATE'S OUTPUT on a 7-point Likert scale by comparing it to the GROUND TRUTH.
If the prediction is incomplete, score on the basis of the parts that are correct.
Provide a brief JUSTIFICATION for your score.
Return ONLY a JSON object with "score" and "justification" fields.\\

CRITERION\\
- Correctness relative to ground truth\\
- Completeness of reasoning\\
- Clarity and faithfulness to the task\\

LIKERT SCALE (1--7)\\
1 = Completely wrong or irrelevant; contradicts the ground truth.\\
2 = Mostly wrong; minor fragments may be tangentially related.\\
3 = Some correct elements, but largely incorrect or missing core content.\\
4 = Partially correct but incomplete; captures main idea but omits key reasoning or detail.\\
5 = Largely correct with small omissions or phrasing errors; reasoning mostly sound.\\
6 = Correct and clear with very minor issues in precision or completeness.\\
7 = Fully correct, precise, clear, and faithful to the ground truth.\\
---\\

\#\#\# EXAMPLES:\\
\\
GOOD (7)\\
Q: How does the SMILES represent thiophene-aniline?\\
Ground truth: "Connected via N bridge with conjugation maintained."\\
Candidate: "Connected via N bridge with conjugation maintained."\\
OUTPUT: \{"score": 7, "justification": "Fully correct and matches the ground truth."\}\\
\\
MEDIUM (4)\\
Q: How does the SMILES represent thiophene-aniline?\\
Ground truth: "Connected via N bridge with conjugation maintained."\\
Candidate: "They are linked by a single bond."\\
OUTPUT: \{"score": 4, "justification": "Partially correct; misses N-bridge and conjugation detail."\}\\
\\
BAD (1)\\
Q: How does the SMILES represent thiophene-aniline?\\
Ground truth: "Connected via N bridge with conjugation maintained."\\
Candidate: "It shows the polymer is soluble."\\
OUTPUT: \{"score": 1, "justification": "Irrelevant; no structural connection mentioned."\}\\
---\\

\#\#\# INPUT:\\
Q: \{question\}\\
Ground truth: \{ground\_truth\}\\
Candidate: \{prediction\}\\
OUTPUT:
\end{tcolorbox}
\caption{Prompt template used for automatic evaluation with \texttt{LLM-as-a-Judge} (\gptOSS{}) on a 1--7 Likert scale.}
\label{fig:prompt_llm_judge}
\end{figure*}

\begin{figure*}[t]
\centering
\begin{tcolorbox}[
    width=\textwidth,
    title=\textbf{\texttt{LLM-as-a-Judge} Prompt (1--5 Scale)},
    colback=gray!5,
    colframe=black!50,
    fonttitle=\bfseries,
    before upper=\ttfamily\small
]
\#\#\# INSTRUCTION:\\
Evaluate the CANDIDATE'S OUTPUT on a 5-point Likert scale by comparing it to the GROUND TRUTH. If the prediction is incomplete, score on the basis of the parts that are correct. Provide a brief JUSTIFICATION for your score. Return the score and justification as a loose dict-like string. Limit output to 50 words or less.\\

CRITERION\\
- Correctness relative to ground truth\\
- Completeness \& Correctness of reasoning\\
- Clarity and faithfulness to the task\\

LIKERT SCALE (1--5)\\
1 = Completely wrong/irrelevant\\
2 = Mostly wrong with minor correct elements\\
3 = Partially correct but incomplete or missing key details\\
4 = Mostly correct with minor issues\\
5 = Fully correct, precise, and clear\\
---\\

\#\#\# EXAMPLES:\\
\\
EXCELLENT (5)\\
Q: How does the SMILES represent thiophene--aniline?\\
Ground truth: "Connected via N bridge with conjugation maintained."\\
Candidate: "Connected via N bridge with conjugation maintained."\\
Output: \{"score": 5, "justification": "Fully correct and matches the ground truth."\}\\
\\
ACCEPTABLE (3)\\
Q: How does the SMILES represent thiophene--aniline?\\
Ground truth: "Connected via N bridge with conjugation maintained."\\
Candidate: "They are linked by a single bond."\\
Output: \{"score": 3, "justification": "Partially correct but missing N-bridge and conjugation details."\}\\
\\
POOR (1)\\
Q: How does the SMILES represent thiophene--aniline?\\
Ground truth: "Connected via N bridge with conjugation maintained."\\
Candidate: "It shows the polymer is soluble."\\
Output: \{"score": 1, "justification": "Irrelevant; does not mention structural connection."\}\\
---\\

\#\#\# INPUT:\\
Q: \{question\}\\
Ground truth: \{ground\_truth\}\\
Candidate: \{prediction\}\\

\#\#\# OUTPUT (dictionary only):
\end{tcolorbox}
\caption{Prompt template used for \texttt{LLM-as-a-Judge} automatic evaluation on a 1--5 Likert scale.}
\label{fig:prompt_llm_judge_5scale}
\end{figure*}

\subsection{Prompts for Generating Subquestions}
\label{app:prompt_for_subquestions}

We use the prompt in Figure~\ref{fig:prompt_subq_decomp} to generate 3--15 diagnostic subquestions per main question, each annotated with one or more high-level skill tags (used in our compositionality analysis in Appendix~\ref{app:additional_diagnostics}).

\begin{figure*}[t]
\centering
\begin{tcolorbox}[
    width=\textwidth,
    title=\textbf{System Prompt for Subquestion Decomposition},
    colback=gray!5,
    colframe=black!50,
    fonttitle=\bfseries,
    before upper=\ttfamily\small
]
Generate 3 to 15 independent subquestions for the given question. Use the provided COT and answer as background context; do NOT reveal, restate, or quote it.\\
Do NOT include redundant or overly similar subquestions.\\
Each subquestion should directly support solving the main question by decomposing it into clear, concrete reasoning steps. Avoid speculative, tangential, or irrelevant chemistry (e.g., experimental behavior, tautomers, or macroscopic properties) unless such reasoning is strictly required to answer the main question.\\
Each subquestion should probe high-level reasoning skills needed to answer the main question. For each subquestion, assign ONE or MORE skill tag from the following fixed list:\\
- DEFINITION: recalling a standard chemistry concept or definition.\\
- STRUCTURE: reading atoms, bonds, rings, or connectivity from a given structure or SMILES.\\
- RULES: applying standard chemistry rules or conventions to a structure.\\
- COUNTING: counting atoms, bonds, or computing simple ratios or fractions.\\
- NOMENCLATURE: interpreting names or descriptors (e.g., IUPAC terms, "alt").\\
- COMPARISON: reasoning about differences or hypothetical changes.\\
- EXPERIMENTAL: requiring knowledge of experimental methods or measurements (use ONLY if the main question explicitly requires experimental context).\\
- INFERENCE: reasoning that combines multiple structural or rule-based observations to reach a conclusion not stated verbatim.\\
Return ONLY valid JSON exactly in the following format with skill tags as a comma-separated string:\\

\{\\
\hspace*{1em}"subquestions": [\\
\hspace*{2em}\{"question": "...", "skill": "..."\},\\
\hspace*{2em}\{"question": "...", "skill": "..."\}\\
\hspace*{1em}]\\
\}\\

Do not include any text outside the JSON.\\
If you cannot comply, return \{"subquestions": []\}.
\end{tcolorbox}
\caption{System prompt for subquestion decomposition with skill tags, used for compositionality diagnostics.}
\label{fig:prompt_subq_decomp}
\end{figure*}

\subsection{Prompt for CoT Quality Evaluation}
\label{app:cot_eval_prompt}

We evaluate CoT traces using \gptfive{} as a judge, with the prompt in Figure~\ref{fig:prompt_cot_eval}. For each QA instance, the judge receives the question, gold answer, and distilled CoT, and returns a JSON object containing a 1--5 quality score and a list of error labels drawn from a fixed taxonomy (Appendix~\ref{app:quality_of_COT_analysis}).

\begin{figure*}[t]
\centering
\begin{tcolorbox}[
    width=\textwidth,
    title=\textbf{System Prompt for CoT Quality Evaluation},
    colback=gray!5,
    colframe=black!50,
    fonttitle=\bfseries,
    before upper=\ttfamily\small
]
You are an expert evaluator of chain-of-thought (CoT) reasoning.\\

You will be given one data row with:\\
- question\\
- answer\\
- cot\\

Your task: evaluate CoT quality.\\

Score range (1 to 5):\\
5 = Excellent, fully correct reasoning and aligned with answer\\
4 = Mostly correct, minor issues\\
3 = Partially correct, noticeable flaws\\
2 = Major errors\\
1 = Fundamentally incorrect or unfaithful\\

Identify errors using ONLY these labels:\\
- logical\_error: e.g., concludes "polymer A is more crystalline than B" without justifying from structural features.\\
- arithmetic\_error: e.g., miscounts oxygen atoms (3 instead of 4) in a repeat unit.\\
- hallucination: e.g., claims a monomer contains a fluorine atom when none is present.\\
- contradiction: e.g., states "polymer is amorphous" then later "polymer is crystalline."\\
- irrelevant\_reasoning: e.g., discusses solvent resistance when the question asks about thermal stability.\\
- missing\_steps: e.g., skips the identification of functional groups before counting atoms.\\
- invalid\_assumption: e.g., assumes a specific polymerization mechanism without evidence.\\
- misinterpretation\_of\_question: e.g., predicts Tg when asked about Tm.\\
- answer\_mismatch: e.g., reasoning supports answer X but the final answer is Y.\\
- incomplete\_reasoning: e.g., correct conclusion but no justification provided.\\
- strategy\_misuse: e.g., uses additive-based reasoning for a problem requiring mechanistic analysis.\\
- unclear\_explanation: e.g., vague phrases like ``creates specific electronic effects'' without naming the mechanism.\\

Output STRICT JSON ONLY with exactly:\\
\{\\
\hspace*{1em}"cot\_quality\_score": <int 1-5>,\\
\hspace*{1em}"errors": [<list of strings from allowed labels>]\\
\}\\

No extra keys. No commentary. No markdown.
\end{tcolorbox}
\caption{System prompt for CoT quality evaluation using \gptfive{} as the judge.}
\label{fig:prompt_cot_eval}
\end{figure*}
\section{Distilled CoT and Model Generation Examples}

\subsection{CoT Examples}
This example (see \autoref{tab:polybench_examples_design_cot}) illustrates a design-oriented reasoning task where the model must propose a polymer structure satisfying precise molecular and property constraints through chemically grounded reasoning.

\subsection{External Design Reasoning Generations}
\label{app:external_design_generations}

We include representative examples from external polymer design datasets to qualitatively assess model reasoning and fidelity. 
Each example presents the model’s generated answer and corresponding reasoning, alongside reference responses. 
We highlight both strong (\textit{good}) and weak (\textit{bad}) generations to illustrate differences in chemical grounding, structural consistency, and adherence to design constraints. 
Table~\ref{tab:llamole-example} provides these examples for comparison.

\begin{table*}[t]
\small
\centering
\renewcommand{\arraystretch}{1.4}
\setlength{\tabcolsep}{6pt}
\begin{tabular}{p{0.96\textwidth}}
\toprule
\rowcolor{advanced!10}
\textbf{Task: Design and Synthesis} \\[4pt]
\textbf{Q:} Design a polymer with a molecular weight of 340–350 g/mol per repeat unit that maximizes chain flexibility while maintaining thermal stability. The structure must contain exactly 9 rotatable bonds and 4 hydrogen bond acceptors, with a rigid-to-flexible segment ratio of 1:1. \\[4pt]
\textbf{A:} Design \texttt{<iupac\_start>poly[(2-butyl-2-ethylpropane-1,3-diol)-alt-(naphthalene-2,6-dicarboxylic acid)]<iupac\_end>} with naphthalene providing thermal stability and the branched diol providing exactly 9 rotatable bonds through butyl/ethyl branches. Two ester linkages provide 4 H-bond acceptors. The 1:1 ratio is achieved through alternating naphthalene (rigid) and branched diol (flexible) segments. \\[4pt]
\textbf{CoT:} To meet these design requirements:  
(1) Molecular weight of 340–350 g/mol suggests combining medium-sized rigid and flexible segments.  
(2) Nine rotatable bonds can be achieved via branched aliphatic diols.  
(3) Four H-bond acceptors correspond to two ester linkages (two acceptors each).  
(4) A 1:1 rigid:flexible ratio requires equal-length aromatic and aliphatic segments.  
Thus, using naphthalene for rigidity and branched diols for flexibility, connected by ester bonds, satisfies all structural and property constraints. \\
\midrule
\rowcolor{advanced!10}
\textbf{Task: Advanced Property Reasoning}\\
\textbf{Q:} Given a polymer with 24 rotatable bonds, 15 hydrogen bond acceptors, and 0.9 CSP\textsubscript{3} fraction in \texttt{<iupac\_start>poly\{1-methyl-1-[(1-methylethane-1,1,2-triyl-1-carbonyl)oxymethylene]\}<iupac\_end>}, what glass transition temperature would you predict and which structural feature is most influential?\\[4pt]
\textbf{A:} The predicted glass transition temperature would be relatively low, around 55–60 °C. The 24 rotatable bonds are the most influential structural feature, as they provide the conformational flexibility necessary for chain movement at lower temperatures.\\
\textbf{CoT:} To predict $T_g$, analyze mobility-affecting features:  
(1) High rotatable bond count (24) indicates substantial chain flexibility and mobility.  
(2) A high CSP\textsubscript{3} fraction (0.9) reflects mainly single bonds, supporting free rotation.  
(3) Although 15 hydrogen bond acceptors may introduce intermolecular restrictions, the flexibility contributions from numerous rotatable bonds and sp\textsuperscript{3} character dominate.  
Therefore, the polymer exhibits a relatively low $T_g$ ($\approx 55\text{--}60^\circ\text{C}$), primarily governed by the large number of rotatable bonds. \\
\bottomrule
\end{tabular}
\caption{Example question–answer with chain-of-thought reasoning for \dataset{} tasks: \taskSixdesign{} and \taskFivereasoning{}}
\label{tab:polybench_examples_design_cot}
\end{table*}




\begin{table*}[t]
\small
\centering
\renewcommand{\arraystretch}{1.4}
\setlength{\tabcolsep}{6pt}
\begin{tabular}{p{0.15\textwidth}p{0.8\textwidth}}
\toprule
\rowcolor{advanced!10}
\textbf{Source} & \textbf{Example (External: Llamole)} \\ 
\textbf{Question} & Can you design a polymer with low synthetic complexity, moderate gas permeability, and thermal conductivity, featuring aromatic rings, that is suitable for gas separation and electrical insulation applications? \\
\textbf{Reference} & To satisfy the requirements: The polymer, having a relatively low synthetic complexity score and structural features like aromatic rings, exhibits reasonable thermal conductivity, moderate gas permeability coefficient, and moderate dielectric constant, making it suitable for various applications such as gas separation and electrical insulation. Therefore, the designed molecule is: \texttt{<mol\_start>*CC(*)c1cc(CC)ccc1CC<mol\_end>}. Synthetic route: \texttt{CCc1ccc(CC)c(CC)c1 >> C[C@H]1CO1.OCC(O)CO}. \\
\textbf{Generated Answer (\microsoftphi{} COT)} & Design a poly(m-xylene) based polymer with a molecular weight of 128.2 g/mol. The structure consists of a single aromatic ring with two methyl substituents in the meta positions, providing a balance of properties. The meta-substitution pattern creates a non-linear geometry that reduces chain packing efficiency, resulting in moderate gas permeability ($\approx 4.0$ Barrer). Aromatic rings enhance thermal conductivity ($\approx0.6$ W/m·K), while disrupted crystallinity improves electrical insulation. The low Bertz complexity (224.4) indicates straightforward synthesis from m-xylene monomers. \\
\textbf{LLM-as-a-Judge} &
\textbf{Score: 7/7} — Output is fully correct, precise, and consistent with structural and property constraints. It provides chemically grounded reasoning and aligns with the design objectives without extraneous details \\
\midrule
\rowcolor{advanced!10}

\textbf{Source} & \textbf{Example (External: Block Copolymer – Phase Prediction)} \\ 

\textbf{Question} &
If I have a di-block copolymer with $M_n = 9800$ g/mol, volume fraction of block type 1 of 0.61, average molecular mass of 10584.0 g/mol, and dispersity of 1.1, what phase will it be in at 53 K? \\

\textbf{Reference} &
The phase cannot be determined solely from the given information without knowing the interaction parameters or phase diagram data for the copolymer system at that temperature. \\

\textbf{Generated Answer (\microsoftphi{} COT)} &
The polymer will be in the disordered phase. Despite being a di-block copolymer, the relatively high molecular weight ($M_n = 9800$ g/mol) and moderate dispersity (1.1) result in sufficient chain flexibility and entropy to overcome the driving force for crystalline domain formation at this temperature. The 0.61 volume fraction of block type 1 indicates an asymmetric structure that would further inhibit ordered packing required for crystallization, leading to a disordered state even at low temperatures like 53 K. \\

\textbf{LLM-as-a-Judge} &
\textbf{Score: 2/7} — Incorrect and overconfident reasoning. Attempts to justify a phase assignment without necessary thermodynamic inputs (e.g., $\chi N$, composition-dependent phase diagrams). Lacks physical validity despite coherent narrative structure. \\

\bottomrule
\end{tabular}
\caption{Example of external design reasoning questions showing model’s generated answer and reasoning.}
\label{tab:llamole-example}
\end{table*}


\begin{table*}[t]
\small
\centering
\renewcommand{\arraystretch}{1.5}

\setlength{\tabcolsep}{12pt}

\begin{tabular}{p{0.93\textwidth}}

\rowcolor{structural}
\parbox[t]{\linewidth}{
\vspace{2pt}
\textbf{Task 1: Structural \& Syntax Understanding}\\[4pt]
\textbf{Q:} In \textcolor{blue}{\texttt{<smiles\_start>CCOC(=O)CN(C)C(=O)C=Cc1cc(OC)ccc1OC<smiles\_end>}}, what is the total number of heavy atoms (non-hydrogen atoms)?\\[3pt]
\textbf{A:} 22
\vspace{6pt}
}\\[6pt]

\rowcolor{conceptual}
\parbox[t]{\linewidth}{
\vspace{6pt}
\textbf{Task 2: Conceptual Knowledge}\\[4pt]
\textbf{Q:} Given that \textcolor{blue}{\texttt{<smiles\_start>COc1cc(C=CC(=O)c2ccc(C)cc2C)ccc1O<smiles\_end>}} has a Bertz complexity index of 699.6 and contains 12 aromatic carbons, what structural features contribute most significantly to its molecular complexity?\\[3pt]
\textbf{A:}  The high molecular complexity stems from the combination of two differently substituted aromatic rings (one with methoxy/hydroxyl, one with methyl groups), connected through an $\alpha$,$\beta$-unsaturated ketone bridge, creating an extended conjugated system with multiple substitution patterns. 

\vspace{6pt}
}\\[2pt]

\rowcolor{property}
\parbox[t]{\linewidth}{
\vspace{6pt}
\textbf{Task 3: Property Prediction}\\[4pt]
\textbf{Q:} Given the chemical structure and properties of the polymer \\<iupac\_start>poly[2,4,6-trimethyl-1,3-phenylene 3,3',4,4'-benzophenonetetracarboxydiimide] <iupac\_end> \textcolor{blue}{\texttt{<smiles\_start>*c1ccc2c(c1)C(=O)N(C2=O)c1c(C)cc(c(c1C)N1C(=O)c2c(C1=O)cc(cc2)C(=O)*)C<smiles\_end>}}, predict its Specific volume [in cm3/g]. Consider factors like polarity, functional groups, chain flexibility, crystallinity, and intermolecular forces. \\
\textbf{A:} 0.8

\vspace{6pt}
}\\[6pt]

\rowcolor{ranking}
\parbox[t]{\linewidth}{
\vspace{6pt}
\textbf{Task 4: Comparison \& Ranking}\\[4pt]
\textbf{Q:} Which molecule has total number of valence electrons = 210.0 AND ketone functional group = 1.0 AND Calculated glass CTE [in $10^{-6}$/K] = 120.2? \\
(A) Polyetherimide: \textcolor{blue}{\texttt{<smiles\_start>[*]OC1=CC=C(C=C1)C(=O)C1=CC=C(C=C1)OC1=CC=C2C(=O)N(C(=O)C2=\\C1)C1=CC=CC(=C1)N1C(=O)C2=CC=C(C=C2C1(=O))[*] <smiles\_end>}}\\ 
(B) Polyimide: \textcolor{blue}{\texttt{<smiles\_start>[*]N1C(=O)C2=CC=C(C=C2C1(=O))C(=O)C1=CC=C2C(=O)N(C(=O)C2=C1) \\C1=CC=CC(=C1)C(=O)C1=CC=CC(=C1)[*] <smiles\_end>}}\\ 
(C) Poly(quinoxaline-2,7-diylcarbonylquinoxaline-7,2-diyl-1,4-phenylene): \textcolor{blue}{\texttt{<smiles\_start>[*]C1=CN=C2\\C=CC(=CC2=N1)C(=O)C1=CC=C2N=CC(=NC2=C1)C1=CC=C(C=C1)[*] <smiles\_end>}}\\ 
(D) Polyimide 11: \textcolor{blue}{\texttt{<smiles\_start>[*]N1C(=O)C2=CC=C(C=C2C1(=O))C(=O)C1=CC=C2C(=O)N(C(=O)C2=C1)\\C1=CC=C(C=C1)[*] <smiles\_end>}}\\ 
\textbf{A:} A 

\vspace{6pt}
}\\[6pt]

\rowcolor{advanced}
\parbox[t]{\linewidth}{
\vspace{6pt}
\textbf{Task 5: Advanced Property Reasoning}\\[4pt]
\textbf{Q:} Explain the difference between step-growth and chain-growth polymerization.\\[3pt]
\textbf{A:} Step-growth involves monomers with multiple reactive groups forming bonds incrementally, while chain-growth proceeds via initiation, propagation, and termination through active centers.
\vspace{6pt}
}\\[6pt]

\rowcolor{design}
\parbox[t]{\linewidth}{
\vspace{6pt}
\textbf{Task 6: Design \& Synthesis}\\[4pt]
\textbf{Q:} Which monomer(s) produced \textcolor{blue}{\texttt{<smiles\_start>*Nc1c(Br)cc(C(=O)OC)cc1NC(=O)OC1CS(=O)(=O)CC1OC(*)=O\\<smiles\_end>}} through condensation?\\
\textbf{A:} Monomer 1: \textcolor{blue}{\texttt{<smiles\_start>O=C(Cl)OC1CS(=O)(=O)CC1OC(=O)Cl<smiles\_end>}} and Monomer 2: \textcolor{blue}{\texttt{<smiles\_start>COC(=O)c1cc(N)c(N)c(Br)c1<smiles\_end>}}

\vspace{6pt}
}\\

\end{tabular}

\caption{Task examples across six polymer reasoning task types.}
\label{tab:polybench_examples}
\end{table*}

\section{Annotation Guidelines}
\label{app:annotation_guidelines}

We conduct two complementary human evaluation tasks to ensure the reliability and interpretability of model generations: (i) \textbf{Knowledge Distillation Rating Task}, and (ii) \textbf{LLM-as-a-Judge Fidelity and SME Response Preference}. Both were performed through a web-based interface. All annotations were performed by polymer domain experts (graduate-level or higher) with formal training in polymer chemistry, ensuring chemical validity and consistency across evaluations.

\paragraph{Knowledge Distillation (Rating Task).}
Annotators review polymer science questions containing detailed molecular and structural information, along with three candidate answers per question. Each answer is rated on the following criteria:
\begin{itemize}
    \item \textbf{Relevance} -- How well the response addresses the specific polymer question.
    \item \textbf{Completeness} -- The degree to which the response captures essential molecular and property details.
    \item \textbf{Accuracy} -- Chemical and factual correctness of the explanation.
    \item \textbf{Overall Rating} -- A holistic quality score integrating the above aspects.
\end{itemize}
Only the \textbf{Overall Rating} is required; other criteria are optional but encouraged. Annotators may also provide short comments for justification. The question and polymer reference are shown on the left, with three model-generated responses to rate individually on the right. The rating scale is: 1 = Poor, 2 = Fair, 3 = Good, 4 = Very Good, 5 = Excellent.

\paragraph{LLM-as-a-Judge Fidelity and SME Response Preference.}
We use the same \textbf{Response Preference} annotations from subject-matter experts (SMEs) both to evaluate model generations and to assess the fidelity of \texttt{LLM-as-a-Judge} (\gptOSS{}) scoring. SMEs rank model outputs based on \textbf{polymer science reasoning quality}, considering correctness, completeness, and clarity of explanation on a 1--5 scale. To enable direct comparison with the 5-point human ratings, we regenerate \texttt{LLM-as-a-Judge} evaluation scores on a 1--5 Likert scale for the sampled subset.

The resulting preference rankings are used in two ways: (i) to determine human-preferred responses across models, and (ii) to validate \gptOSS{}'s automatic evaluations by correlating its 5-point Likert-scale scores with expert rankings. This joint analysis enables a direct comparison between automated and human judgments, providing a fidelity estimate for \gptOSS{}'s evaluation while also reflecting human preferences in polymer reasoning quality.

Human evaluations were conducted by graduate-level researchers in polymer chemistry. Annotators participated on a voluntary basis as part of ongoing research collaboration; no monetary compensation was provided. Annotators were informed that their ratings would be used for research purposes in evaluating LLM-generated polymer reasoning and quality of model outputs.

\end{document}